\documentclass{article}
\PassOptionsToPackage{numbers, compress}{natbib}

\usepackage[preprint]{neurips_2026}
\usepackage[table]{xcolor}
\usepackage{wrapfig}
\usepackage[utf8]{inputenc} % allow utf-8 input
\usepackage[T1]{fontenc}    % use 8-bit T1 fonts
\usepackage{hyperref}       % hyperlinks
\usepackage{url}            % simple URL typesetting
\usepackage{booktabs}       % professional-quality tables
\usepackage{amsfonts}       % blackboard math symbols
\usepackage{wrapfig} 
\usepackage{amsmath}
\usepackage{amsthm}
\usepackage{nicefrac}       % compact symbols for 1/2, etc.
\usepackage{microtype}      % microtypography
\usepackage{xcolor}         % colors
\usepackage{amssymb} 
\usepackage{makecell}
\usepackage{graphicx}
\usepackage{multirow}

\usepackage{algorithm}
\usepackage{algorithmic}
\usepackage{subcaption}
\title{Compositional Concepts Erasing: Disentangling Object-Style via Domain Manifold Unlearning}

\author{%
  \normalfont Yang Xiao$^{1}$, Kaiyuan Deng$^{2}$, Gen Li$^{3}$, Zinan Ling$^{1}$\\
  Zhiwen Fan$^{4}$, Xiaolong Ma$^{2}$, Bo Hui$^{1}$\thanks{Corresponding Author: bo-hui@utulsa.edu}\\
  $^{1}$University of Tulsa \quad $^{2}$The University of Arizona\\
  $^{3}$Clemson University \quad $^{4}$Texas A\&M University
}

\begin{document}

\maketitle

\begin{abstract}
Diffusion models pose significant risks regarding the generation of illicit content and copyright violations. This has catalyzed the development of machine unlearning to erase harmful concepts or copyrighted styles from a model’s generative repertoire without compromising its general performance. However, previous unlearning approaches on diffusion models have largely focused on either object-level or style-level removal and failed to address the challenge of object-style combination unlearning. At the same time, multiple unlearning requests often lead to catastrophic forgetting. To address these challenges, we propose \textbf{Domain Manifold Unlearning (DMU)}, which uses an EMA-updated feature proxy extracted from intermediate diffusion features to represent the retained target-object direction.
% which operates directly on the manifold through a non-parametric statistic updated with an exponential moving average (EMA). 
As a lightweight estimating representation of an object, DMU enables selective unlearning of domain-specific components in a specific object while preserving other semantic representations, thereby significantly enhancing robustness in multiple combination unlearning scenarios. Extensive experiments show that DMU significantly mitigates catastrophic forgetting, outperforming state-of-the-art methods of batch unlearning and sequential unlearning.
\end{abstract}

\vspace{-3mm}
\section{Introduction}
\vspace{-3mm}
With the rapid advancement of artificial intelligence, the expanding capabilities of foundation models, such as diffusion models, have raised significant concerns regarding data privacy protection, copyright compliance, and safe deployment~\cite{cooper2024machine,li2025machine}.  For example, users can craft prompts that manipulate these models into producing unethical outputs, such as violent imagery, sexually explicit content, or protected copyrighted material. Driven by regulations, such as the “right to be forgotten”, laws against NSFW and copyright infringement by generative AI~\cite{regulation2016regulation}, machine unlearning has rapidly emerged as a critical pillar~\cite{li2025sculpting,deng2026forget,xiao2025efficient,xiao2025knowledge,xiao2025right,kawamura2025approximate,zhang2024unlearncanvas,lee2025continual,thakral2025continual}. This paradigm enables models to selectively erase the influence of specific data or concepts without requiring complete retraining from scratch.

While previous unlearning approaches on diffusion models have largely focused on either object-level or style-level removal, a more challenging setting is to forget \textbf{object-style combinations}~\cite{kawamura2025approximate,zhang2024unlearncanvas} (e.g., a human in Van Gogh style). Such fine-grained unlearning provides more flexibility and reliability in real applications.
For example, in content moderation systems, a model may need to forget specific object–style combinations, such as "weapons" depicted in cartoon styles, while still retaining the ability to recognize weapons in real-world images and benign cartoon content. Removing only the object or the style would lead to over-suppression, highlighting the need for fine-grained unlearning of object–style combinations.
% For example, a model in an autonomous driving system should forget ``humans" in animation styles in case humans in movie posters distract driving. 
As another example, recent controversies around AI-generated images in the style of specific studios or artists (e.g., Studio Ghibli-like renderings) highlight a realistic need for fine-grained control over object–style combinations. In practice, a platform may permit generating images of a public figure or character in generic styles, while restricting the generation of the same object in the distinctive style of a protected studio, franchise, or living artist. This requires selectively removing only the prohibited subject–style combination, rather than eliminating the entire object or the entire style. While pre-trained diffusion models exhibit strong generalization across diverse datasets, they often entangle semantic and stylistic representations within the latent space. As a result, existing object-level or style-level unlearning approaches are insufficient to satisfy such fine-grained and selective unlearning requirements.

\begin{figure}
    \centering
    \vspace{-2mm}
    \includegraphics[width=0.85\linewidth]{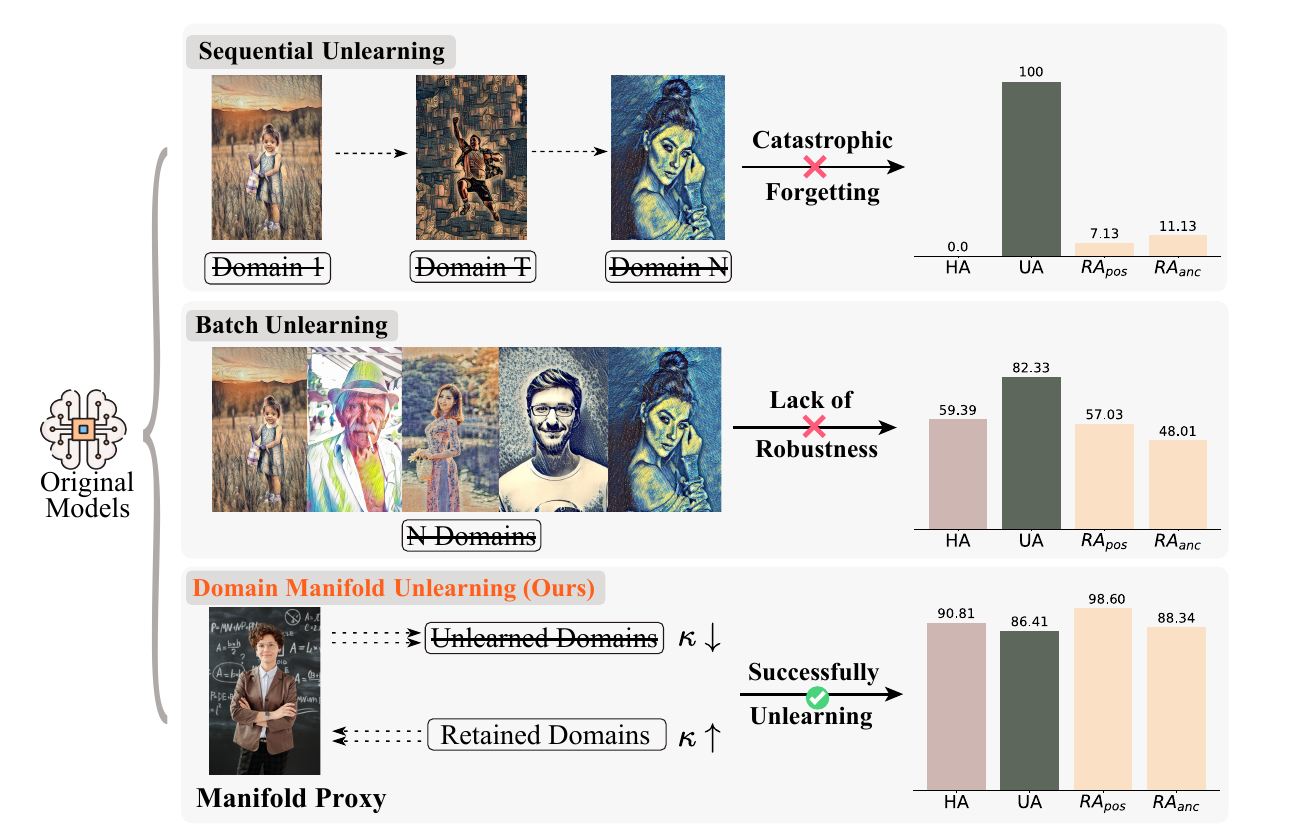}
    \caption{Motivation of our method. HA refers to the harmonic average of performance across the later three metrics: UA refers to 100 minus the accuracy in the negative set (Unlearned domains. RA$_{pos}$ refers to accuracy in the positive set (Retained domains). RA$_{anc}$ refers to accuracy in the anchor set (Other objects in all styles).}
    \label{fig:pipeline}
    \vspace{-6mm}
\end{figure}

Existing unlearning methods can be broadly categorized into batch unlearning (BU) and sequential unlearning (SU)~\cite{deng2026forget,deng2026forget2, kawamura2025approximate,liu2022right,tarun2023fast,li2025sculpting}. BU removes multiple concepts simultaneously in a single request, while SU processes removal requests incrementally over time. However, neither setting is well suited for fine-grained object–style combination unlearning. 
There is a formidable challenge of BU: because the unlearning gradients associated with different domain distributions may interfere with one another, a simple linear superposition often results in incomplete unlearning or an excessive degradation of the model's general utility. Studies have demonstrated that independently designed unlearning operations can clash, causing the model to exhibit extreme instability when processing "hard-to-unlearn" samples that span multiple domains~\cite{liu2022right, zhang2024unlearncanvas}.
SU~\cite{zhang2024unlearncanvas,lee2025continual} addresses scenarios in which domain unlearning requests arrive in an asynchronous and continuous manner. In this setting, the model must incrementally remove the influence of specific data distributions, such as particular illustration styles, while strictly preserving its diagnostic and generative capabilities on the retained data.
Our preliminary studies have consistently observed a significant decline in performance on retained data as the number of unlearning requests increases.  As shown in Figure~\ref{fig:pipeline}, the retaining accuracy of target object (in the remaining style) and other objects dropped catastrophically. This degradation cannot be attributed solely to simple changes in model weights~\cite{lee2025continual}, rather, it arises from gradient noise introduced during earlier unlearning operations, which is progressively amplified in subsequent tasks. In the deeply entangled latent space of foundation models, forcefully erasing multiple domains using conventional gradient descent often irreparably damages the underlying representation structure.

% Another challenge is Catastrophic Forgetting in Batch Unlearning (BU) and Sequential Unlearning (SU). \textbf{Batch Unlearning}~\cite{deng2026forget,deng2026forget2, kawamura2025approximate,liu2022right,tarun2023fast,li2025sculpting} requires a model to simultaneously remove multiple domains within a single request. 

To overcome these fundamental limitations, it is essential to move beyond treating multi-target domain unlearning as a simple linear optimization problem. Recognizing that domain characteristics and core semantic knowledge are intricately entangled within a low-dimensional manifold structure in the model's parameter space, we propose \textbf{Domain Manifold Unlearning} (DMU). Instead of directly applying conflicting or noisy gradients, our method explicitly models the underlying manifold of the target domains. By navigating and decoupling the domain-specific manifolds from the shared semantic representation, DMU effectively neutralizes the gradient conflicts inherent in Batch Unlearning and mitigates the noise amplification in Sequential Unlearning. This approach allows the model to precisely excise the targeted domain distributions while comprehensively preserving its core utility and generalization capabilities. 
We summarize our contributions as follows:
\begin{itemize}
    \item Unlike prior methods that only focus on either object or style-level removal, we propose a framework to address object-style combination unlearning.
    \item Our approach significantly enhances robustness in multiple unlearning requests by directly operating on the domain manifold, providing a more precise way to isolate targeted data.
    \item Through extensive experiments, we demonstrate that DMU outperforms state-of-the-art methods of batch unlearning and sequential unlearning.
\end{itemize}

\section{Method}
\vspace{-3mm}
\subsection{Preliminaries}
\vspace{-3mm}

\begin{figure}
    \centering
    \includegraphics[width=1.00\linewidth]{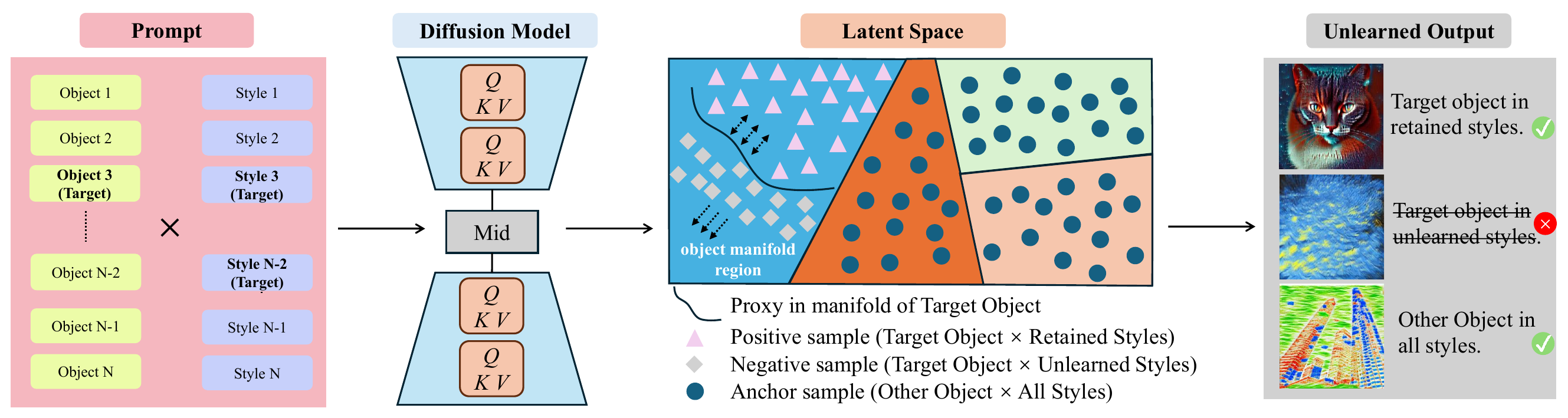}
    \caption{Method}
    \label{fig:method}
\end{figure}

% To achieve unlearning while preserving model
% performance, the unlearning problem is formulated as an optimization task to update parameters from their pretrained weights. Let the $\mathcal{D}_{u}$ refers to the targeted unlearn data, $\mathcal{D}_{r}$ refers to the retained data. Here, we define the objective function as:
% \begin{equation}
%     \mathcal{L}=\lambda\mathcal{L}_{u}(\theta, \mathcal{D}_{u})+\mathcal{L}_{r}(\theta, \mathcal{D}_{r})
% \end{equation}

% Where $\mathcal{L}_{u}$ refers to the unlearn method to erase the impact of some data or concepts, and $\mathcal{L}_{r}$ refers to the retain loss to keep the model's performance on retained data.

% In our setting, we divide the dataset into three components: 1) Positive Set: $\mathcal{D}_{t, r}$ refers to the data of target classes on retained styles. 2) Negative Set: $\mathcal{D}_{t, u}$ refers to the data of target classes on unlearned styles. 3)Anchor Set:  $\mathcal{D}_{r}$ refers to the data of other classes on all styles:

% \begin{equation}\label{eq:vanillaloss}
%     \mathcal{L}=\lambda\mathcal{L}_{u}(\theta, \mathcal{D}_{t, u},\mathcal{D}_{t, r})+\mathcal{L}_{r}(\theta, \mathcal{D}\setminus \left(\mathcal{D}_{t, u} \right))
% \end{equation}

To achieve unlearning while preserving model
performance, the unlearning problem is formulated as an optimization task to update parameters from their pretrained weights. In our setting, we divide the dataset into three components: 1) Positive Set: $\mathcal{D}_{t, r}$ refers to the data of target object on retained styles. 2) Negative Set: $\mathcal{D}_{t, u}$ refers to the data of target object on unlearned styles. 3) Anchor Set:  $\mathcal{D}_{r}$ refers to the data of other objects on all styles:

\vspace{-5mm}
\begin{equation}\label{eq:vanillaloss}
    \mathcal{L}=\lambda\mathcal{L}_{u}(\theta, \mathcal{D}_{t, u},\mathcal{D}_{t, r})+\mathcal{L}_{r}(\theta, \mathcal{D}\setminus \left(\mathcal{D}_{t, u} \right))
\end{equation}

\vspace{-5mm}
\subsection{Non-parametric Object Proxy in the Manifold}
To implement the unlearning objective $\mathcal{L}_{u}$ effectively, we introduce a non-parametric object proxy. As illustrated in Figure~\ref{fig:method}, given a batch of object--style pairs, our goal is to remove the target object under specific styles while preserving its representation under retained styles. In diffusion models, we extract the feature vector $\mathbf{h}(x)$ from intermediate representations of the U-Net (e.g., bottleneck layers) during the denoising process. We then use these features to approximate an object-specific proxy via dynamically updating. 

A key challenge is that, in general, we cannot guarantee that each mini-batch contains sufficient positive samples (i.e., target object under retained styles, $x \in \mathcal{D}_{t,r}$). Directly using per-batch statistics would therefore lead to unstable or even undefined estimates of the proxy. To address this, we adopt an exponential moving average (EMA) to accumulate information across batches, ensuring a stable and consistent estimation.

To further avoid contamination of domains to be unlearned, the EMA is updated exclusively using positive samples. Let $i$ denote the current optimization step. The proxy is updated as:
\begin{equation}\label{eq:EMA}
    \boldsymbol{\mu}^{(i)} = m\cdot\boldsymbol{\mu}^{(i-1)} + (1-m)\cdot \mathbb{E}_{x \in \mathcal{D}_{t,r}}[\mathbf{h}^{(i)}(x)], \quad \boldsymbol{\mu}^{(0)}=\mathbf{0}
\end{equation}

In the special case where each mini-batch is guaranteed to contain sufficient positive samples, the proxy can be estimated directly from the current batch (i.e., setting momentum $m=0$). In our ablation study, we empirically verify that this simplified variant achieves comparable performance under controlled sampling, suggesting that EMA mainly serves as a stabilizing mechanism when batch statistics are unreliable.

With random data shuffling and sufficient training iterations, it provides a stable approximation of the feature expectation. This manifold therefore acts as a semantic reference in the representation space. To formally define our objective, we first $\ell_2$-normalize the feature and the proxy:

\vspace{-5mm}
\begin{equation}
\tilde{\mathbf{f}}(x)=\frac{\mathbf{h}(x)}{\|\mathbf{h}(x)\|} \in\mathbb{S}^{d-1},\qquad
\tilde{\boldsymbol{\mu}}=\frac{{\boldsymbol{\mu}}}{\|{\boldsymbol{\mu}}\|} \in\mathbb{S}^{d-1}
\end{equation}
where $d$ refers to the dimension of the feature (In Stable Diffusion, it is 1280).

\subsection{The view of von Mises-Fisher (vMF) distribution}
\paragraph{Connection to the vMF distribution.}
We can interpret $\tilde{\mathbf{f}}(x)$ as directional data modeled by a vMF distribution on $\mathbb{S}^{d-1}$:
\begin{equation}
p(\tilde{\mathbf{f}}\mid \tilde{\boldsymbol{\mu}},\kappa)=C_d(\kappa)\exp\!\left(\kappa\,\tilde{\boldsymbol{\mu}}^\top \tilde{\mathbf{f}}\right),
\end{equation}
Here, $\kappa$ describes the concentration around the proxy $\tilde{\boldsymbol{\mu}}$. Therefore, maximizing the vMF likelihood is fundamentally equivalent to maximizing the cosine similarity $\tilde{\boldsymbol{\mu}}^\top \tilde{\mathbf{f}}$.

Accordingly, our manifold loss can be viewed as a likelihood-based objective: positive samples $\mathcal{D}_{t,r}$ are encouraged to align with $\tilde{\boldsymbol{\mu}}$ (high likelihood under vMF with $\tilde{\boldsymbol{\mu}}$), while negative samples $\mathcal{D}_{t,u}$ are pushed away from $\tilde{\boldsymbol{\mu}}$ (equivalently, higher likelihood under a vMF with the $-\tilde{\boldsymbol{\mu}}$), thereby preserving retained semantics and disrupting the targeted object semantics under unlearned styles.

In practice, to formulate our empirical loss, we directly optimize the geometric direction, omitting the explicit dynamic scalar $\kappa$ from the objective function. This leads to the following DMU loss:
\begin{equation}
\mathcal{L}_{\mathrm{u}}(x)=
\begin{cases}
    -\tilde{\boldsymbol{\mu}}^\top \tilde{\mathbf{f}}(x), & x\in\mathcal{D}_{t,r}\\[3pt]
    \ \ \tilde{\boldsymbol{\mu}}^\top \tilde{\mathbf{f}}(x), & x\in\mathcal{D}_{t,u}
\end{cases}
\end{equation}

% \vspace{-5mm}

\begin{wraptable}{r}{0.33\columnwidth}
    \centering
    \caption{
    % Evolution of vMF $\kappa$ on DMU. 
    The larger the $\kappa$, the more it aligns with the object's semantics; the smaller the value, the further it diverges.}
    \vspace{-3mm}
    \begin{tabular}{c|c|c|}
    \toprule
    \midrule
    Steps & $\kappa_{pos} $ &  $\kappa_{neg}$  \\
    \midrule
    0 & 1815.85 & 1773.03 \\
    500 & 1843.38 & 637.38\\
    1000 & 4576.56 & 0 \\
    \midrule
    \bottomrule
    \end{tabular}
    \label{tab:kappa}
    \vspace{-5mm}
\end{wraptable}

\paragraph{Overall Objective}
While the DMU loss $\mathcal{L}_{\mathrm{u}}$ is responsible for decoupling the domain-specific features of the target object, we must also ensure that the model does not suffer from catastrophic forgetting of other general concepts. To achieve this, we maintain the standard diffusion denoising loss, denoted as $\mathcal{L}_{\text{r}}$. The original denoising loss is applied exclusively to the positive set $\mathcal{D}_{t,r}$ and the anchor set $\mathcal{D}_{r}$ to preserve core generative semantics. The final overall optimization objective is a combination of the knowledge preservation loss and the domain unlearning manifold loss:
\begin{equation}
    \mathcal{L} = \mathbb{E}_{x \in \mathcal{D}_{t,r} \cup \mathcal{D}_{r}} \left[ \mathcal{L}_{\text{r}}(x) \right] + \lambda \mathbb{E}_{x \in \mathcal{D}_{t,r} \cup \mathcal{D}_{t,u}} \left[ \mathcal{L}_{\mathrm{u}}(x) \right]
\end{equation}
where $\lambda$ is a hyperparameter (Empirically, $0.05 \sim 1.00$ works for unlearning, See in Table~\ref{tab:abla_lambdas}). 

This formulation ensures that the domain unlearning operations remain highly targeted without interfering with the shared semantic knowledge space.

\vspace{-3mm}
\paragraph{Dynamic Evolution of vMF Concentration.}
This dynamic evolution is mathematically grounded in the properties of the vMF distribution. Specifically, the expected cosine similarity is given by:
\begin{equation}\label{eq:kappa}
    \mathbb{E}[\tilde{\boldsymbol{\mu}}^\top \tilde{\mathbf{f}}] = A_d(\kappa) = \frac{I_{d/2}(\kappa)}{I_{d/2-1}(\kappa)}
\end{equation}
where $I_v(\cdot)$ denotes the modified Bessel function of the first kind at order $v$. Crucially, $A_d(\kappa)$ is a strictly monotonically increasing function of $\kappa$. Therefore, explicitly optimizing $\mathcal{L}_{\mathrm{u}}$ strictly corresponds to manipulating the semantic concentration $\kappa$ of the feature space:
\begin{itemize}
    \item \textbf{Positive alignment ($\kappa_r \uparrow$):} Minimizing the loss for positive samples increases their cosine similarity, thereby implicitly increasing their concentration $\kappa_r$ around $\tilde{\boldsymbol{\mu}}$. It ensures the core semantics of the target object remain robust.
    \item \textbf{Negative repulsion($\kappa_u \downarrow$):} Pushing negative samples away from $\tilde{\boldsymbol{\mu}}$ continuously decreases their expected similarity, dropping their concentration $\kappa_u$. As $\kappa_u \to 0$, the feature distribution degenerates toward a uniform distribution on the hypersphere. This geometrically explains the process of ``losing semantics'', where the target object features in unlearned styles are completely dispersed and structurally destroyed.
\end{itemize}

Simultaneously, the original denoising loss $\mathcal{L}_{\text{r}}$ acts as a crucial spatial regularizer. While the proxy inherently encourages positive samples to converge towards a single center (increasing $\kappa_r$), $\mathcal{L}_{\text{r}}$ counterbalances this by preserving the rich intra-class variance. It ensures that positive samples maintain their domain-specific stylistic differences rather than experiencing dimensional collapse, thereby preserving the original generative diversity.

\paragraph{Empirical estimation of concentration.}
To quantitatively analyze how concentrated the retained and unlearned features are around the shared proxy in the manifold, 
we further estimate the vMF concentration parameter from the extracted feature embeddings. Given normalized features $\tilde{\mathbf{f}}_j \in \mathbb{S}^{d-1}$ and a normalized shared proxy $\tilde{\boldsymbol{\mu}}$, we first compute the empirical mean cosine similarity
\begin{equation}
\bar{R} = \frac{1}{N}\sum_{j=1}^{N} \tilde{\boldsymbol{\mu}}^\top \tilde{\mathbf{f}}_j,
\end{equation}
which serves as an empirical estimate of the mean resultant length under the vMF model. 

Since $A_d(\kappa)$ is strictly increasing but does not admit a simple closed-form inverse (See in Equation~\ref{eq:kappa}), we adopt the standard high-dimensional approximation to estimate $\kappa$ from $\bar{R}$:
% \begin{equation}
% \hat{\kappa} \approx \frac{\bar{R}(d-\bar{R}^2)}{1-\bar{R}^2}.
% \end{equation}
\begin{equation}
\hat{\kappa} =
\begin{cases}
0, & \bar{R}<0, \\
\dfrac{\bar{R}(d-\bar{R}^2)}{1-\bar{R}^2}, & \bar{R}\ge 0.
\end{cases}
\end{equation}
In our setting, the feature dimension is high ($d=1280$), making this approximation accurate and numerically stable. Intuitively, a larger $\bar{R}$ corresponds to a larger $\hat{\kappa}$, indicating that features are more tightly concentrated around the proxy; conversely, a smaller $\bar{R}$ implies a weaker concentration and a more dispersed feature distribution.

\begin{table}[]
    \centering
    \caption{Results over different momentum.}
    \begin{tabular}{c|c|c|c|c|c|c|c|c|c|}
    \toprule
    \midrule
        Momentum & 0.1 & 0.2 & 0.3 & 0.4 & 0.5 & 0.6 & 0.7 & 0.8 & 0.9 \\
    \midrule
        HA (↑) & 88.54 & 88.25 & 87.93 & 85.67 & 88.05 & 88.06 & 84.37 & 83.18 & 90.45\\
    \midrule
    \bottomrule
    \end{tabular}
    \label{tab:momentum}
\end{table}

\vspace{-3mm}
\section{Experiments}
\vspace{-3mm}
\subsection{Experiment Setups}
\vspace{-3mm}
We implemented our method under two unlearning scenarios: 1) Sequential Unlearning (SU), where requests are processed chronologically, requiring the model to unlearn the first domain before proceeding to the next in a continuous stream; and 2) Batch Unlearning (BU), where multiple unlearning requests are consolidated and executed simultaneously in a single optimization step. 

\textbf{Goal}: Our objective is to cause the generation of the target object to fail within specific domains, while maintaining the performance of other objects in all domains.

\textbf{Metrics}: Following UnlearnCanvas~\cite{zhang2024unlearncanvas}, we use the followings as metrics: \textbf{UA} is $100\%$ minus the classification accuracy of target object in $D_{t, u}$. \textbf{RA}$_{pos}$ is classification accuracy of target object in $D_{t, r}$. \textbf{RA}$_{anc}$ is classification accuracy of other objects in $D_{r}$. \textbf{HA}~\cite{kawamura2025approximate} is the harmonic average of \textbf{UA}, \textbf{RA}$_{pos}$, and \textbf{RA}$_{anc}$, measuring the balance between unlearning effectiveness and knowledge retention: $\textbf{HA} = 3\times(\frac{1}{\textbf{UA}+1e-8} + \frac{1}{\textbf{RA}_{pos}+1e-8} + \frac{1}{\textbf{RA}_{anc}+1e-8})^{-1}$. \textbf{StyleAcc}$_{R}$ is the classification accuracy of styles in $\mathcal{D}\setminus \mathcal{D}_{t, u}$. \textbf{CLIP}$_{R}$ refers to the clip score of (image, prompt) pairs in $\mathcal{D}\setminus \mathcal{D}_{t, u}$.

% HA's formula:
% \begin{equation}
%      \textbf{HA} = \frac{3}{\frac{1}{\textbf{UA}+1e-8} + \frac{1}{\textbf{RA}_{pos}+1e-8} + \frac{1}{\textbf{RA}_{anc}+1e-8}}
% \end{equation}
% \begin{equation}
%      \textbf{HA} = 3(\frac{1}{\textbf{UA}+1e-8} + \frac{1}{\textbf{RA}_{pos}+1e-8} + \frac{1}{\textbf{RA}_{anc}+1e-8})^{-1}
% \end{equation}

\textbf{Feature Extraction Details}: To construct the proxy, we extract the intermediate feature $\mathbf{h}_t(x)$ from the U-Net's bottleneck. Specifically, we hook the output of the first block within the Middle Block (middle\_block[1]), as it captures the most compressed semantic and domain representations. Since styles are generally independent of spatial layouts, we apply Global Average Pooling over the spatial dimensions to obtain a 1D feature vector, followed by $l_2$-normalization.

\textbf{Datasets \& Models}: The experiments are primarily implemented in UnlearnCanvas~\cite{zhang2024unlearncanvas} benchmark where it has 20 objects and 51 styles, each object-style pair contains 20 samples. Based on this benchmark, we extended our experiments to OfficeHome~\cite{venkateswara2017deep} (65 objects, 4 styles, and 15,500 samples) and PACS~\cite{li2017deeper} (7 objects, 4 styles, and 9991 samples). The experiments are conducted on the SD1.5, SD2, and SDXL.

\definecolor{lightPink}{HTML}{FFB6C1}
\definecolor{HAcolor}{RGB}{255,242,204}  % soft gold
\definecolor{UAcolor}{RGB}{253,237,236}  % soft red

\vspace{-3mm}
\subsection{Results}
\vspace{-2mm}
\subsubsection{Main Experiments}
\begin{table}[]
    \centering
    \caption{Main Experiments. The default request is to unlearn 6 domains.}
    \resizebox{\columnwidth}{!}{
        \begin{tabular}{l||>{\columncolor{HAcolor}}c|>{\columncolor{UAcolor}}c|c|c||c||c}
        \toprule
        \midrule
        Method & HA (↑) & UA(↑) & RA$_{pos}$(↑) & RA$_{anc}$ (↑) & StyleAcc$_{R}$(↑) & CLIP$_{R}$(↑) \\
        \midrule
        FT w/o $\mathcal{D}_{t,u}$ (BU) & 47.98 & 30.24 & 53.33 & 93.40 & 91.27 & 29.10\\
        GA (BU) & 3.31 & 97.78 & 1.13 & 97.73 & 94.16 & 27.56 \\
        L1 (BU) & 69.74 & 48.89 & 82.49 & 95.77 & 94.29 & 28.12\\
        L2 (BU) & 60.38 & 36.44 & 87.22 & 92.76 & 95.17 & 28.83 \\
        \midrule
        % FMN (BU)~\cite{zhang2024forget} & 0 & 100 & 0 & 0 & 2.61 & 22.39\\
        UCE (BU)~\cite{gandikota2024unified} & 59.39 & 82.33 & 57.03 & 48.01 & 94.69& 27.25\\
        SculpMemory (BU)~\cite{li2025sculpting} & 53.63 & 52.09 & 39.69 & 86.62 & 96.23& 27.79\\
        ADU (BU)~\cite{kawamura2025approximate} & 53.63& 52.09&	39.69&	86.62& 89.31& 27.77\\
        SSDA (BU)~\cite{basak2024forget} & 64.08 & 41.47 &	86.93 & 89.27 & 97.12& 29.06\\
        DMU (BU)& \textbf{87.33} & 87.07 & 89.36 & 85.63 & 95.20& 29.08\\
        \midrule
        ESD (SU)~\cite{gandikota2023erasing} & 22.11 & 100 & 11.91 & 23.97 & 93.96& 22.81\\
        EDiff (SU)~\cite{wu2025erasing} & 22.99& 99.17& 14.76& 19.00 & 24.85& 22.68\\
        UCE (SU)~\cite{gandikota2024unified} & 27.95 & 100 & 21.77 & 19.46 & 76.53& 22.94\\
        CUIG (SU)~\cite{lee2025continual} &	 70.49& 56.25& 72.11& 91.64& 97.09& 29.67\\
        DMU (SU) & \textbf{85.81}& 91.13& 84.24& 82.52& 88.90& 29.55\\
        \midrule
        \bottomrule
    \end{tabular}
    }
    \label{tab:MainExperiment}
\end{table}

Table~\ref{tab:MainExperiment} presents the main experimental results under both Sequential Unlearning (SU) and Batch Unlearning (BU) settings on the UnlearnCanvas benchmark, where six domains are removed by default (see details in Appendix). \textbf{Overall}, \textbf{DMU} consistently achieves the \textbf{SOTA} HA across both settings. Compared to existing methods, it not only significantly improves UA, but also preserves higher retained accuracy on both positive samples ($\text{RA}_{pos}$) and anchor samples ($\text{RA}_{anc}$), while maintaining competitive style consistency and semantic alignment (StyleAcc$_R$ and CLIP$_R$). 

We use the HA to characterize the trade-off effects of unlearning; a low HA corresponds to: 1) unlearning failure, or 2) successful unlearning that lacks robustness. In the SU setting, prior methods like ESD, EDiff, and UCE achieve nearly perfect UA scores, but suffer from severe degradation in retained performance, with $\text{RA}_{pos}$ and $\text{RA}_{anc}$ dropping significantly. In contrast, CUIG improves retention performance but still yields relatively limited unlearning effectiveness. DMU effectively mitigates this trade-off, achieving a high HA and UA, while maintaining strong retained performance. These results indicate that DMU can remove target knowledge without excessively damaging the retained distribution. In the BU setting, a similar trend is observed. Existing methods either prioritize retention (e.g., SSDA achieves high $\text{RA}_{pos}$ but low UA) or emphasize unlearning at the cost of performance degradation. DMU again achieves the best overall performance, with the highest HA and UA, while maintaining competitive retained accuracy and generation quality. Notably, DMU outperforms all baselines in jointly optimizing unlearning effectiveness and retention, demonstrating its robustness under BU setting.

\begin{table}[]
    \centering
    \caption{Other Datasets.}
    \resizebox{\columnwidth}{!}{
        \begin{tabular}{l||c||>{\columncolor{HAcolor}}c|>{\columncolor{UAcolor}}c|c|c||c||c}
            \toprule
            \midrule
            Dataset & Method & HA (↑) & UA(↑) & RA$_{pos}$(↑) & RA$_{anc}$ (↑) & StyleAcc$_{R}$(↑) & CLIP$_{R}$(↑) \\
            \midrule
            \multirow{3}{*}{OfficeHome} 
            & ADU & 56.61 & 44.12 & 62.17 & 70.22 & 84.67 & 25.49\\
            & SSDA & 50.77 & 37.50 & 50.00 & 80.47 & 80.19 & 25.17 \\
            & DMU & \textbf{81.35} & 82.35 & 76.47 & 85.78 & 93.26 & 26.41 \\
            \midrule
            \multirow{3}{*}{PACS} 
            & ADU & 58.32& 59.80& 41.18& 95.83& 94.43& 24.39\\
            & SSDA & 64.93& 39.22& 94.12& 99.17& 96.73& 26.57 \\
            & DMU & \textbf{91.63}& 98.04& 88.24& 89.23& 91.49& 25.79 \\
            \midrule
            \bottomrule
        \end{tabular}
    }
    \label{tab:OtherDatasetExperiment}
    \vspace{-6mm}
\end{table}

\begin{figure*}[t]
    \centering
    \begin{subfigure}[t]{0.30\textwidth}
        \centering
        \includegraphics[width=\linewidth]{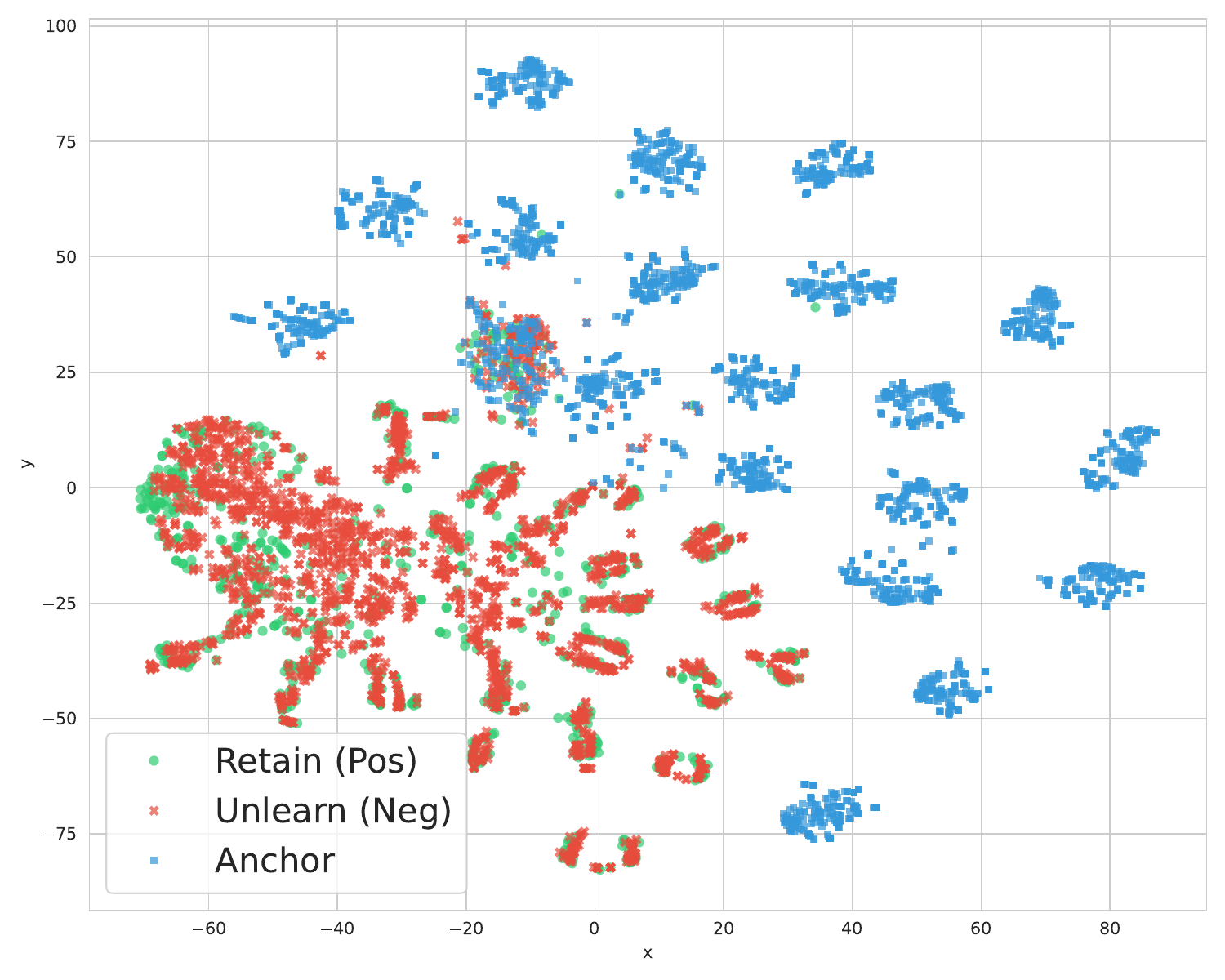}
        \label{fig:gla}
    \end{subfigure}
    \hfill
    \begin{subfigure}[t]{0.30\textwidth}
        \centering
        \includegraphics[width=\linewidth]{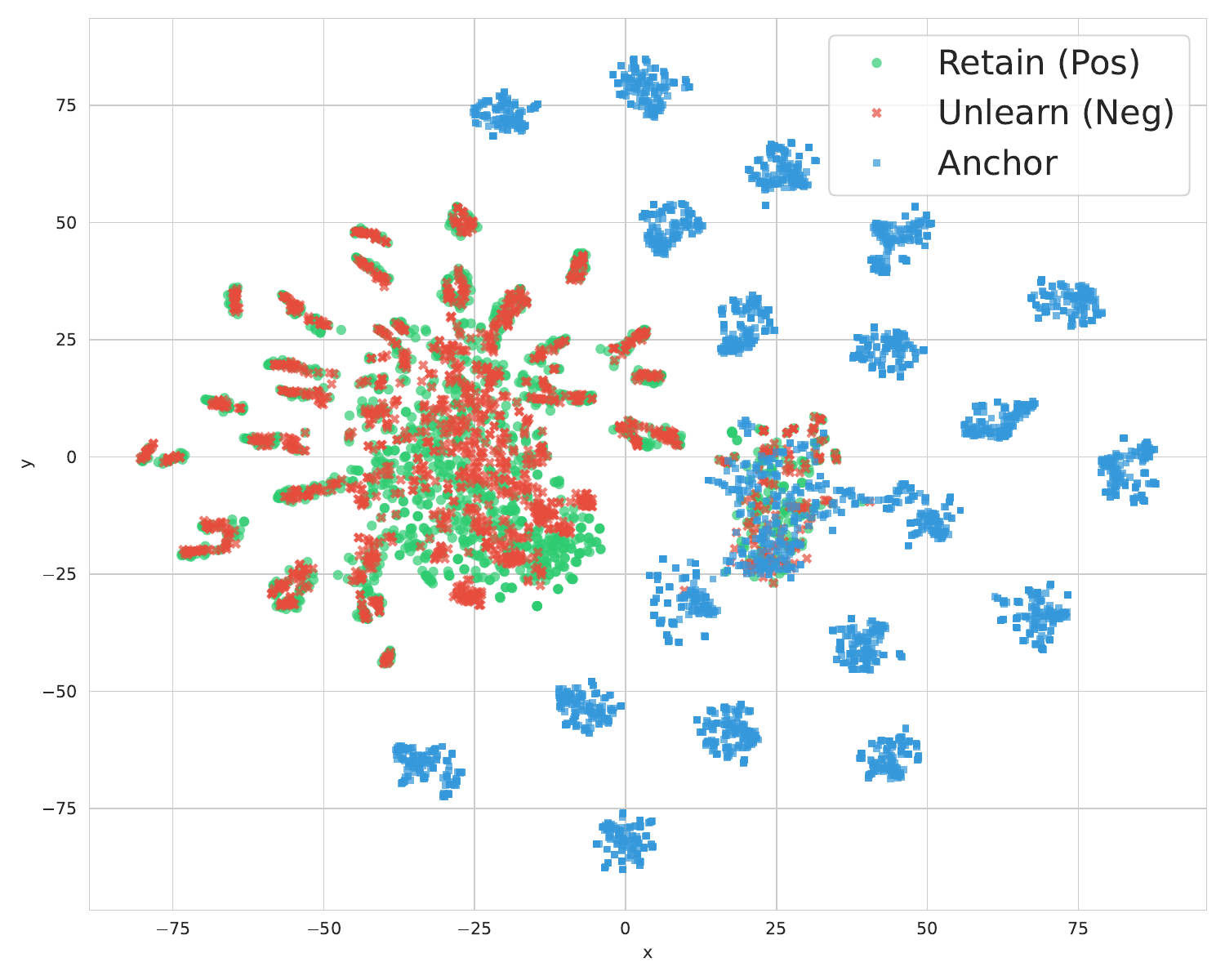}
        \label{fig:sculpting}
    \end{subfigure}
    \hfill
    \begin{subfigure}[t]{0.30\textwidth}
        \centering
        \includegraphics[width=\linewidth]{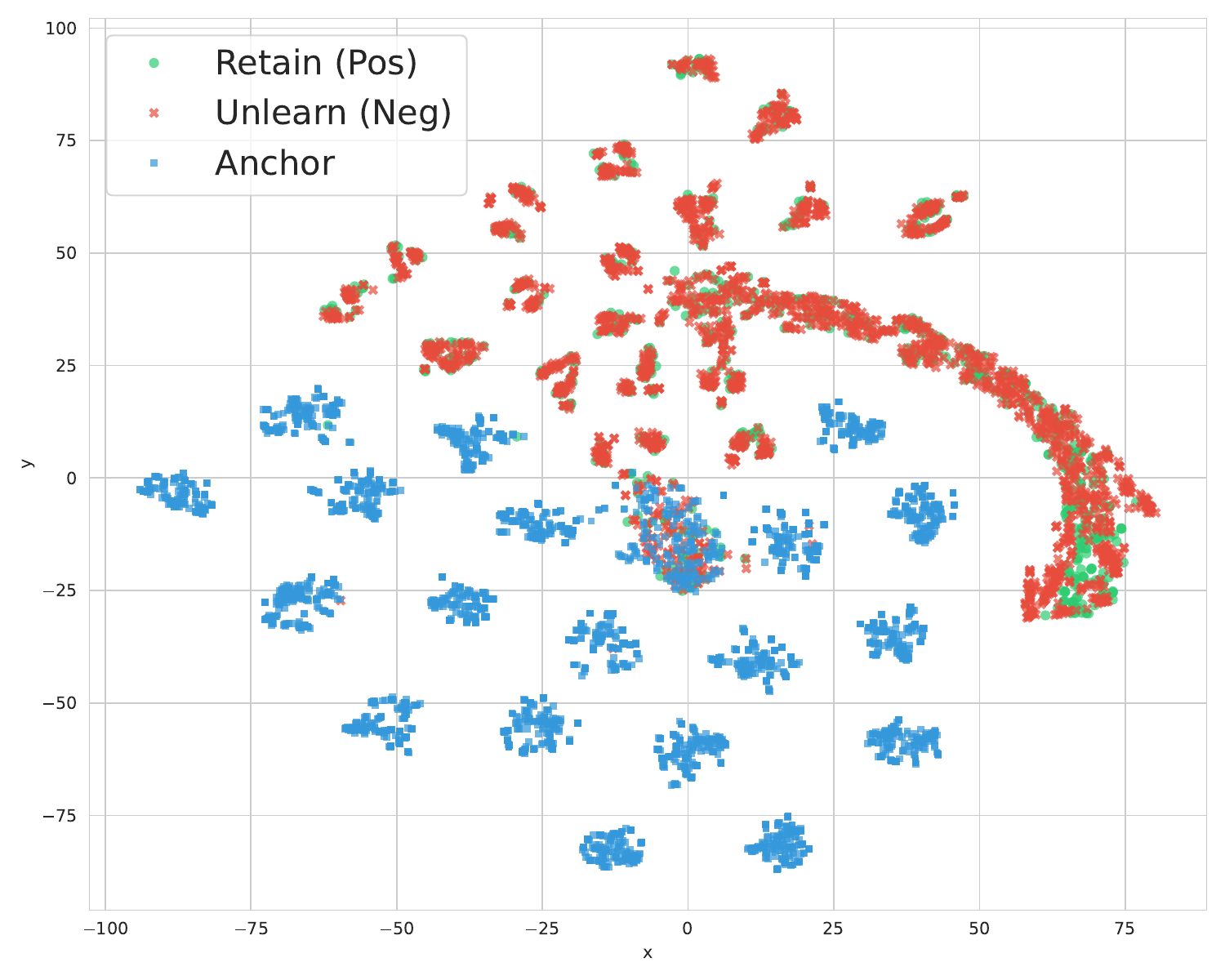}
        \label{fig:MMD}
    \end{subfigure}
    \caption{Manifold Visualization of SSDA, SculptingMemory, and ADU, in 1000 steps.}
    \label{fig:manifold_compare}
\end{figure*}
\begin{figure*}[t]
    \centering
    \begin{subfigure}[t]{0.30\textwidth}
        \centering
        \includegraphics[width=\linewidth]{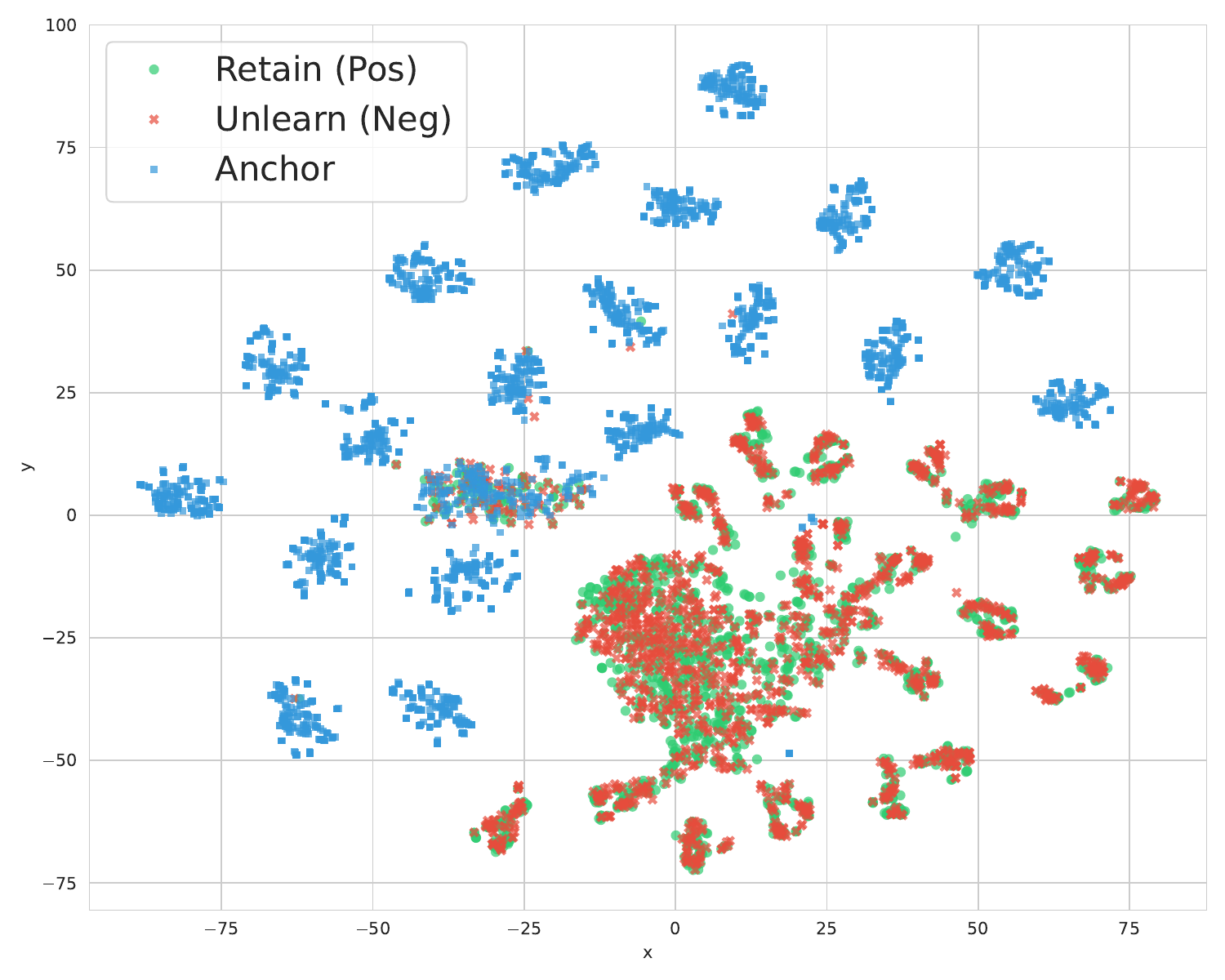}
        \label{fig:step0}
    \end{subfigure}
    \hfill
    \begin{subfigure}[t]{0.30\textwidth}
        \centering
        \includegraphics[width=\linewidth]{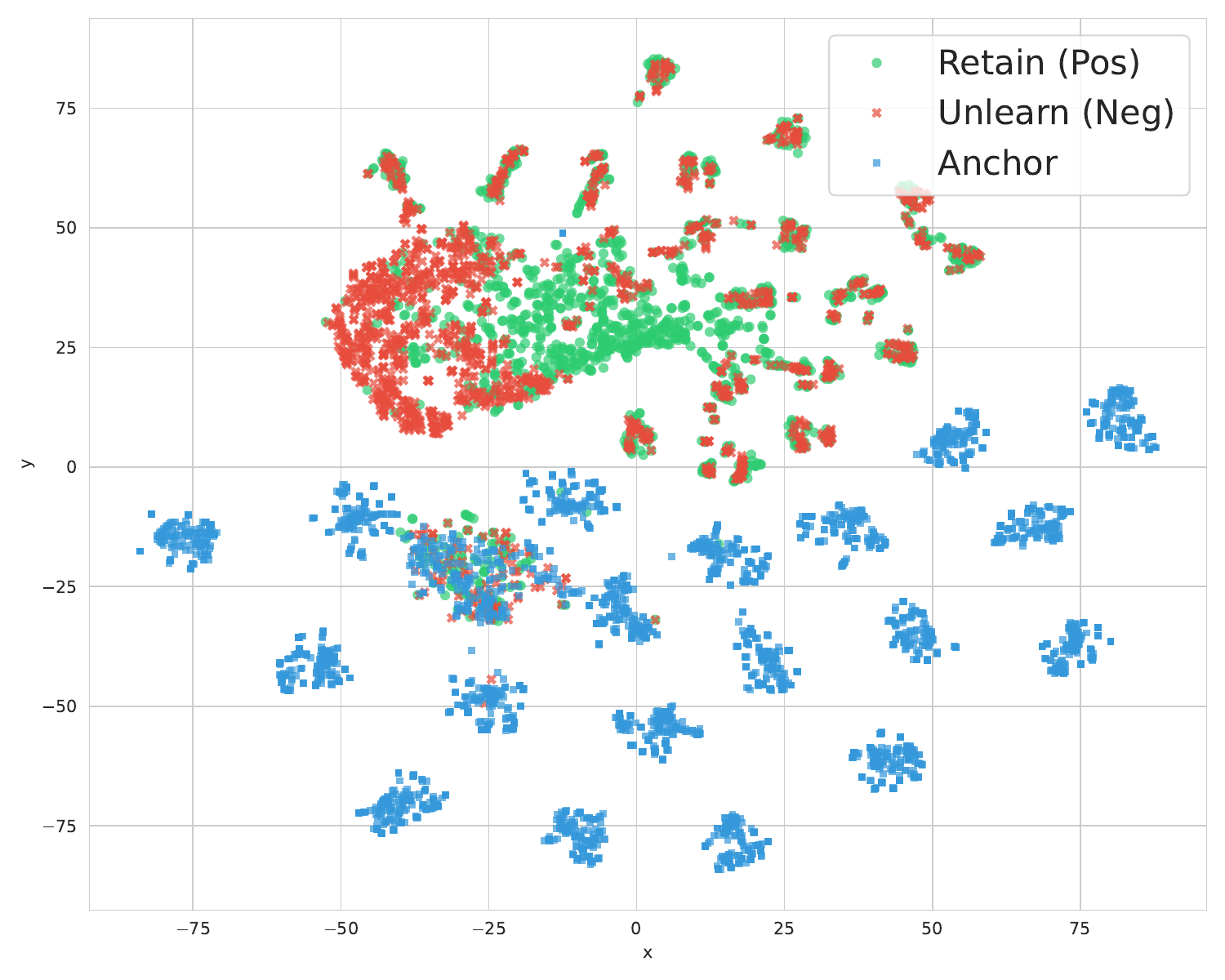}
        \label{fig:step500}
    \end{subfigure}
    \hfill
    \begin{subfigure}[t]{0.30\textwidth}
        \centering
        \includegraphics[width=\linewidth]{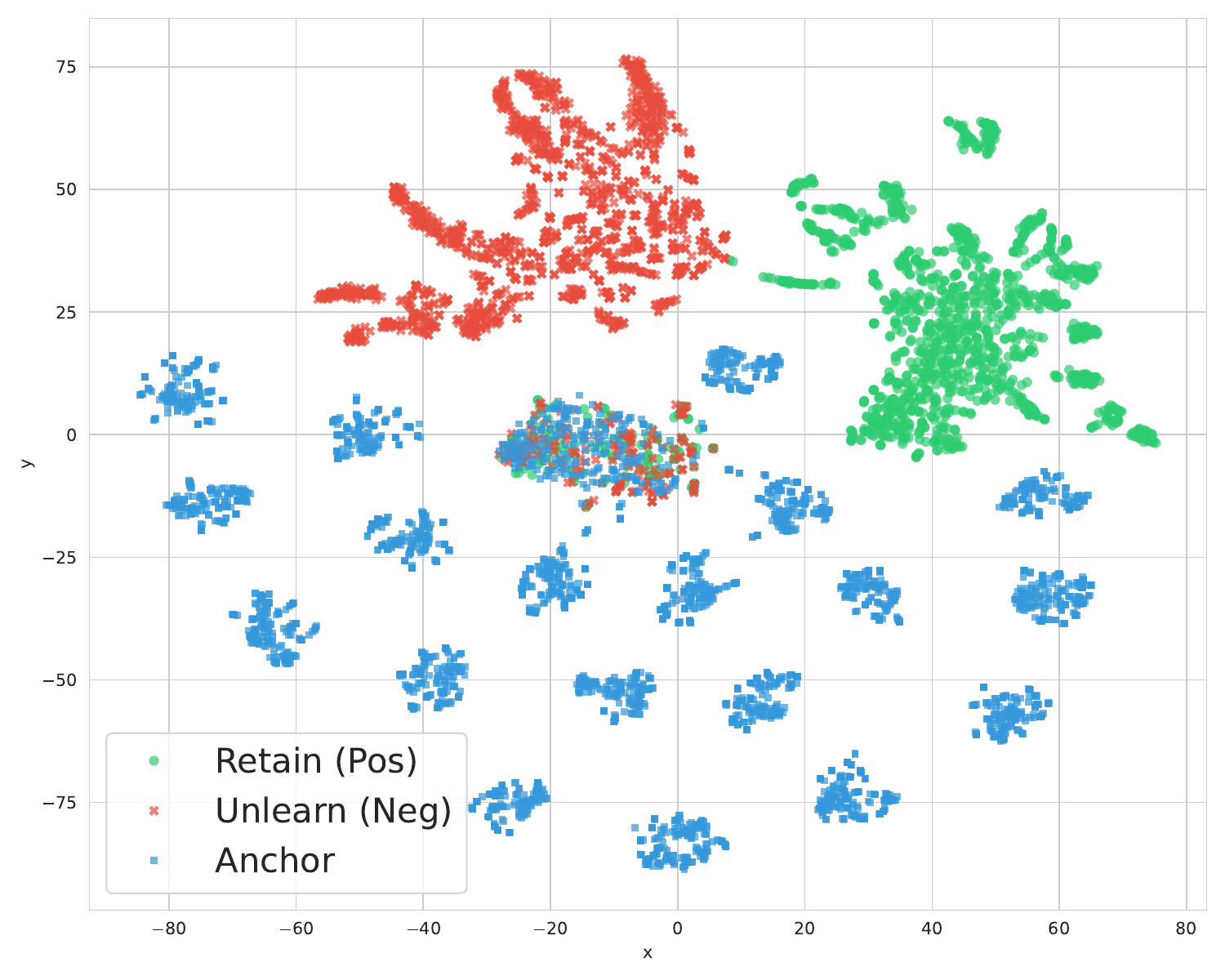}
        \label{fig:step1000}
    \end{subfigure}
    \caption{Manifold Visualization of DMU in 0, 500, and 1000 steps.}
    \label{fig:manifold_vis}
    \vspace{-5mm}
\end{figure*}

% \begin{figure}
%     \centering
%     \includegraphics[width=1.0\linewidth]{Figures/losslandscape.png}
%     \caption{DMU Loss Landscape in 0, 500, and 1000 steps.}
%     \label{fig:Loss_Landscape}
%     \vspace{-3mm}
% \end{figure}

% \begin{figure*}[t]
%     \centering
%     \begin{subfigure}[t]{0.30\textwidth}
%         \centering
%         \includegraphics[width=\linewidth]{Figures/manifold_visualization_Step0.svg}
%         \label{fig:step0}
%     \end{subfigure}
%     \hfill
%     \begin{subfigure}[t]{0.30\textwidth}
%         \centering
%         \includegraphics[width=\linewidth]{Figures/manifold_visualization_Step500.svg}
%         \label{fig:step500}
%     \end{subfigure}
%     \hfill
%     \begin{subfigure}[t]{0.30\textwidth}
%         \centering
%         \includegraphics[width=\linewidth]{Figures/manifold_visualization_Step1000.svg}
%         \label{fig:step1000}
%     \end{subfigure}
%     \caption{Manifold Visualization of DMU in 0, 500, and 1000 steps.}
%     \label{fig:manifold_vis}
%     \vspace{-5mm}
% \end{figure*}

% \begin{figure}
%     \centering
%     \includegraphics[width=1.0\linewidth]{Figures/losslandscape.png}
%     \caption{DMU Loss Landscape in 0, 500, and 1000 steps.}
%     \label{fig:Loss_Landscape}
%     \vspace{-3mm}
% \end{figure}

\begin{table}[]
    \centering
    \caption{Performance of SDXL and SD2 in UnlearnCanvas Dataset.}
    \resizebox{0.85\columnwidth}{!}{
        \begin{tabular}{l||>{\columncolor{HAcolor}}c|>{\columncolor{UAcolor}}c|c|c||c||c}
            \toprule
            \midrule
            Method & HA (↑) & UA(↑) & RA$_{pos}$(↑) & RA$_{anc}$ (↑) & StyleAcc$_{R}$(↑) & CLIP$_{R}$(↑) \\
            \midrule
            SDXL & 6.79& 2.38& 92.7& 89.68 & 97.84 & 28.55\\
            + DMU & \textbf{95.75}& 100.00& 98.10& 89.78 & 97.78 & 28.59 \\
            \midrule
            SD2 & 7.32 & 2.57 & 95.56 & 97.11 & 96.25 & 29.16 \\
            + DMU & \textbf{96.60} & 95.71 & 96.19 & 97.94 & 94.63 & 29.67 \\
            \midrule
            \bottomrule
        \end{tabular}
    }
    \label{tab:SDXLExperiment}
\end{table}

\textbf{Generalization across models and datasets.} Beyond the UnlearnCanvas benchmark, we further evaluate DMU across different model architectures and data distributions. Specifically, we extend our experiments to larger diffusion backbones (SDXL and SD2) on UnlearnCanvas (See in Table~\ref{tab:SDXLExperiment}), and to domain generalization benchmarks (OfficeHome and PACS) based on SD1.5 (See in Table~\ref{tab:OtherDatasetExperiment}). Across all these settings, DMU consistently maintains strong unlearning effectiveness and retention performance, demonstrating that our method is not tied to a specific model scale or dataset, but generalizes well across architectures and domains.

\vspace{-2.5mm}
\subsubsection{Visualization}
Using the unlearned weights, we extracted features for 100 batches from the bottleneck layers of U-Net. To highlight the differences between the positive set and the negative set, we then randomly removed a portion of the anchor samples so that the anchor samples constituted 50\% of the total samples, while the positive and negative samples each accounted for 25\%. Figure~\ref{fig:manifold_compare} shows that three compared methods did not completely decouple the positive set and the negative set. This also explains why the results for \textbf{UA} and \textbf{RA}$_{pos}$ are not robust. Figure~\ref{fig:manifold_vis} shows the decoupling process. 
At the beginning of the unlearning process (step 0), the positive and negative samples are tightly clustered and overlapping in the latent space, indicating that the model retains a strong influence from the negative data. As the process reaches step 500, a noticeable divergence begins to emerge, with the two sets showing initial signs of separation. By the completion of unlearning at step 1000, the positive and negative samples are entirely decoupled and form distinct clusters. This progression demonstrates that the model has successfully eliminated the influence of the negative samples, achieving effective domain unlearning. Table~\ref{tab:kappa} shows the increasing $\kappa_{pos}$ and decreasing $\kappa_{neg}$, which indicates progressive semantic concentration of positive samples and effective dispersion of negative samples.

% \textbf{Loss Landsacpe} At initialization (step 0), the DMU loss landscape is nearly flat. This is caused by the zero-initialized proxy vectors. Since the DMU loss relies on interactions between features and proxies, a zero proxy results in nearly constant similarity scores, making the loss insensitive to parameter perturbations. At step 500, the proxy vectors have moved away from zero, enabling meaningful interactions between features and proxies. As a result, the loss landscape develops a smooth basin, indicating that the model has entered a regime where optimization is guided by informative gradients. At step 1000, the loss landscape becomes more structured and non-convex, reflecting the successful completion of the unlearning objective. At this stage, the model has effectively separated target (to-be-unlearned) concepts from retained ones, leading to a task-aligned minimum. The increased complexity of the landscape indicates that the proxy vectors have encoded rich interactions between positive, negative, and anchor samples.

\subsection{Ablation Studies}
\textbf{Ablation of Loss} As shown in Table~\ref{tab:abla_loss}, the full DMU achieves the best overall balance with the highest HA. Removing \textit{positive alignment} still maintains relatively high UA, but causes RA${pos}$ to drop sharply, showing severe degradation on the preserved positive set. In contrast, removing \textit{negative repulsion} leads to a collapse of UA, indicating failed unlearning despite high retention performance. Removing \textit{denoising preservation} preserves strong UA, but slightly degrades both RA${pos}$, RA$_{anc}$, and StyleAcc$_R$, suggesting its role in stabilizing retained semantics. Finally, jointly removing \textit{positive alignment} and positive-set \textit{denoising preservation} causes RA${pos}$ to further collapse, demonstrating that both objectives are essential for preventing catastrophic forgetting on the preserved target object.

\begin{table}[]
    \centering
    \caption{Performance of Loss Ablation Studies}
    \label{tab:abla_loss}
    \resizebox{\columnwidth}{!}{
        \begin{tabular}{l||>{\columncolor{HAcolor}}c|>{\columncolor{UAcolor}}c|c|c||c||c}
            \toprule
            \midrule
            Method & HA (↑) & UA(↑) & RA$_{pos}$(↑) & RA$_{anc}$ (↑) & StyleAcc$_{R}$(↑) & CLIP$_{R}$(↑) \\
            \midrule
            Full DMU (*) & 87.33 & 87.07 & 89.36 & 85.63 & 95.20& 29.08\\
            * Remove Pos. alignment & 58.77 & 72.92 & 38.15 & 89.89 & 93.61 & 28.76\\
            * Remove Neg. repulsion & 57.81 & 35.42 & 77.46 & 92.98 & 96.73 & 29.19\\
            * Remove Denoising preservation (Pos. \& Anchor set) & 84.39 & 89.58 & 79.48 & 84.73 & 83.80 & 28.79\\
            * Remove Pos. alignment \& Denoising preservation (Pos. set)& 48.07 & 79.17 & 26.01 & 88.24 & 81.30 & 28.70 \\
            \midrule
            \bottomrule
        \end{tabular}
    }
\end{table}
% \vspace{-10mm}
\textbf{Momentum} In this part, we show the impact of the momentum parameter. The momentum of a manifold refers to its tendency to maintain a value in each update of each batch. A momentum value of 0 means the proxy only relies on the current batch features, while a value of 1 results in a static proxy (e.g., remaining at initialization), both of which fail to capture the desired dynamic update behavior of the proxy and are therefore excluded from our analysis. HA exhibits oscillatory behavior when momentum is between 0.1 and 0.9. The overall performance remains robust, with no significant decline observed 
(See in Table~\ref{tab:momentum}).
% (See in Table~\ref{tab:momentum} and Figure~\ref{fig:ha_momentum}).

% \begin{figure*}[t]
%     \centering
%     \begin{subfigure}[t]{0.45\textwidth}
%         \centering
%         \includegraphics[width=\linewidth]{Figures/HA_w_momentum.png}
%         \caption{HA under different momentum.}
%         \label{fig:ha_momentum}
%     \end{subfigure}
%     \hfill
%     \begin{subfigure}[t]{0.45\textwidth}
%         \centering
%         \includegraphics[width=\linewidth]{Figures/radar_request.png}
%         \caption{Results under more unlearning requests.}
%         \label{fig:radar_request}
%     \end{subfigure}
%     \caption{Performance of ablation studies}
%     \label{fig:ablation_studies}
% \end{figure*}

\begin{wrapfigure}{r}{0.40\textwidth}
    \centering
    \vspace{-10mm}
        \includegraphics[width=\linewidth,trim={0 80 0 70},clip]{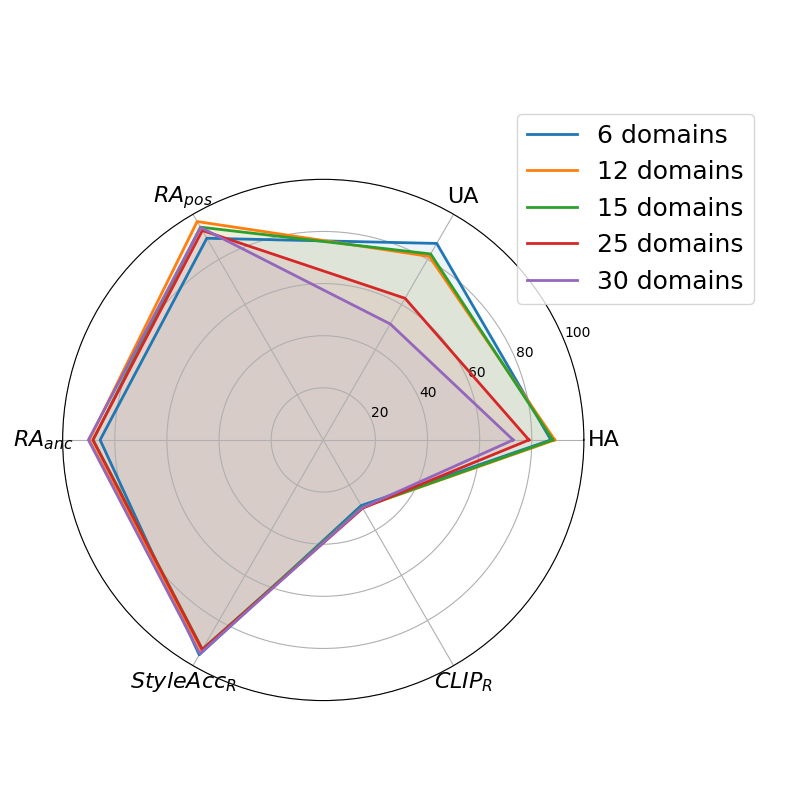}
        \caption{Exp. of Unlearning requests.}
        \label{fig:radar_request}
\end{wrapfigure}
\textbf{More Unlearn Pairs}
In this part, we test the robustness of DMU by increasing the number of unlearning requests. Under the same training timestep (timestep = 1000), we observe that as the number of requests increases, UA gradually decreases, indicating that the unlearning task becomes more challenging with more domains to be removed (See in Figure.~\ref{fig:radar_request}). As the requests increase, the positive set becomes smaller while the anchor set remains unchanged; thus, aside from UA and HA, the metrics show little change. To further investigate this effect, we increase the training timestep to 2000 for the cases with 25 and 30 requests. We find that UA recovers to approximately 85\%, demonstrating that the performance degradation is not due to a limitation of the method itself, but rather insufficient optimization steps. This suggests that as the number of unlearning requests grows, DMU requires more training iterations to effectively decouple unlearned domains' information.

\vspace{-5mm}
\section{Discussion and Limitation}
\vspace{-3mm}
\subsection{Extension to multi-domain unlearning.}
\vspace{-3mm}
In our current experiments, we perform unlearning on a single target object with multiple styles, and report averaged results across different objects. Each experiment maintains an object-specific proxy, which is updated independently using only the corresponding positive samples and does not involve gradient-based optimization. This design naturally enables a more general multi-domain setting, where multiple objects and multiple styles can be jointly unlearned within a unified framework. Specifically, one can construct multiple positive and negative sets corresponding to different object--style combinations, allowing for flexible configurations such as jointly unlearning $N$ objects across $M$ styles, or even heterogeneous settings where each object is associated with a different subset of styles. Due to space limitations, we leave a comprehensive exploration of such multi-domain unlearning scenarios for future work. We believe this direction is particularly promising for scaling unlearning methods to more realistic and complex data distributions.
\begin{figure}[t]
    \centering
    \includegraphics[width=0.75\linewidth]{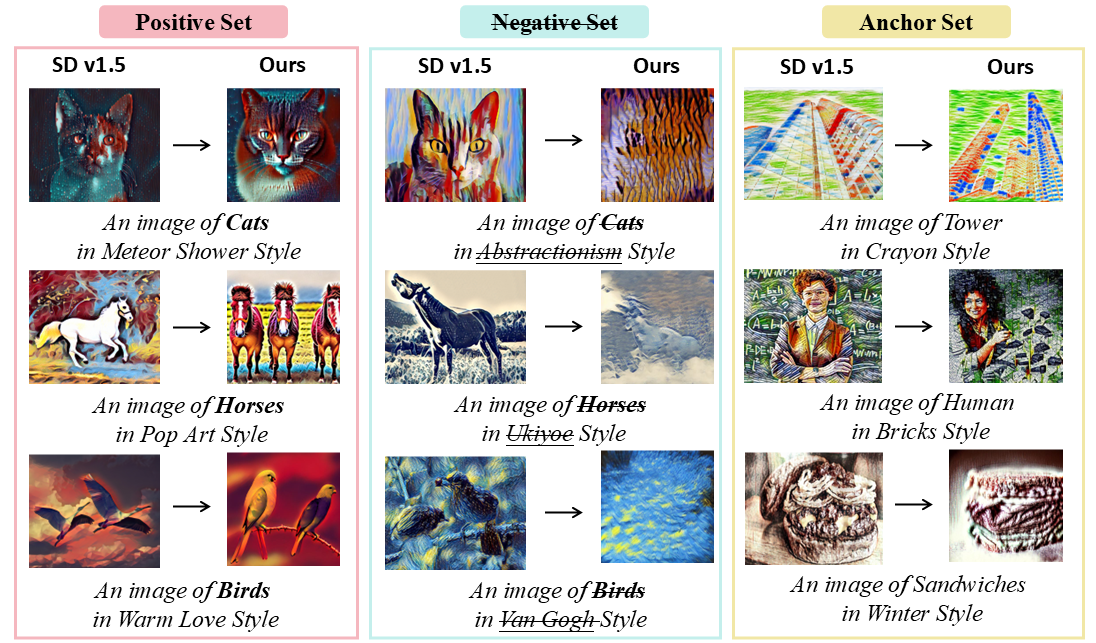}
    \caption{Case figure shows that our method selectively erases the domains (coupling of object-style pairs) while preserving retained domains and maintaining performance on other objects.}
    \label{fig:case}
\end{figure}

\vspace{-3mm}
\subsection{Auditing and Concept Identifiability}
\vspace{-3mm}
Our evaluation measures forgetting and retention under the tested object--style requests; it does not establish resistance to latent-space recovery or concealment of erased concepts. Latent-space auditing~\cite{deng2026no} and erasure-footprint identification~\cite{deng2026your} provide complementary tests that remain to be investigated for DMU.

\vspace{-3mm}
\subsection{Feature Extraction in Different Layers}
\vspace{-3mm}
Due to page constraints and the substantial time cost of our experiments, we do not explore the impact of using different layers for feature extraction in the DMU process. In practice, many diffusion-based works commonly extract features from intermediate (middle) layers~\cite{lan2025diffusion,chang2024sparsedit,lee2025v}, as these representations tend to balance low-level details and high-level semantics. Early layers primarily capture local textures and noise patterns, while deeper layers encode more abstract, task-specific concepts; intermediate layers therefore provide a more stable and informative representation for downstream objectives such as unlearning.

\vspace{-5mm}
\section{Related Works}
\vspace{-3mm}
\subsection{Multiple Concept Erasing}
\vspace{-2mm}
Early concept erasure methods in diffusion models mainly focus on removing a single concept through static fine-tuning or prompt-level editing~\cite{gandikota2024unified,gandikota2023erasing,fan2023salun,lu2024mace,zhao2024separable}. However, real-world unlearning requests often involve multiple concepts that may be semantically correlated or partially overlapping, making direct extension of single-concept methods prone to interference and utility degradation. 
ESD~\cite{gandikota2023erasing} fine-tunes diffusion model weights with negative guidance. 
UCE~\cite{gandikota2024unified} offers a unified closed-form framework for multi-concept forgetting. 
MACE~\cite{lu2024mace} employs LoRA modules alongside closed-form cross-attention refinement to erase a large number of concepts while maintaining the model’s generative performance for other concepts.
SepME~\cite{zhao2024separable} achieves independent, non-interfering removal of multiple concepts by generating concept-irrelevant representations and decoupling weight updates for each concept.
Sculpting Memory~\cite{li2025sculpting} leverages dynamic masks and concept-aware optimization for stable multi-concept erasing. 
Forget-It-All (FIA)~\cite{deng2026forget2} performs training-free multi-concept unlearning by identifying concept-sensitive neurons and fusing their masks while preserving neurons that support general generation. 
CUIG~\cite{lee2025continual} proposes a continual concept unlearning framework for diffusion models that mitigates utility degradation by stabilizing parameter updates and reducing cross-concept interference during sequential removal. 
Extending multi-concept erasure to video generation, FADE~\cite{li2026fade} combines joint closed-form key/value editing with frame-aware low-rank adapters and prompt-dependent routing to address residual concept leakage in individual frames.
These approaches mark progress toward scalable and unified multi-concept erasing, but are still insufficient for large-scale forgetting, especially when erasing (object, style) group.

\vspace{-3mm}
\subsection{Auditing and Identifying Erased Concepts}
\vspace{-3mm}
Beyond erasure effectiveness on standard prompts, recent work examines what unlearned models still reveal or encode. AVCE~\cite{deng2026no} uses latent-space auditing to expose recoverable erased concepts and incorporates auditing-aware updates to strengthen erasure. Tracer~\cite{deng2026your} addresses a complementary question: it identifies which concepts were erased and estimates their number through spectral analysis of weight footprints and limited forward probes. These works motivate evaluating compositional unlearning beyond output-based forgetting and retention metrics.

\vspace{-3mm}
\subsection{Domain Unlearning}
\vspace{-3mm}
The essence of domain unlearning lies in eliminating the statistical footprints of a specific domain from the parameter distribution of a pre-trained model~\cite{kawamura2025approximate, zhang2024unlearncanvas}. 
ADU~\cite{kawamura2025approximate} uses Maximum Mean Discrepancy (MMD) to explicitly disentangle domain distributions and adaptively captures instance-specific domain information. 
SSDA~\cite{basak2024forget} minimizes the classification loss for retained images and maximizes uncertainty loss for unlearned images, then uses Gaussian-guided latent alignment (GLA) strategy that uses a prior distribution to align domain-agnostic information and retained domain information. However, these methods only consider the explicit decoupling of the unlearned and retained domains, they fail to account for whether the retained domain still persists within the manifold of general knowledge. Consequently, the unlearning performance often lacks robustness.

\vspace{-6mm}
\section{Conclusion}
\vspace{-3mm}
\textbf{Main Findings}: In this paper, we present Domain Manifold Unlearning (DMU), a method designed to address the challenges of robustness and catastrophic forgetting in multiple domains unlearning for diffusion models. By explicitly modeling data manifold using a non-parametric proxy updated via EMA, DMU enables the precise decoupling of domain-specific features from shared semantic space. Through the lens of the von Mises-Fisher distribution, we provide a mathematical foundation for the unlearning process, showing that optimizing the manifold loss corresponds to manipulating semantic concentration and dispersion in the latent space. By combining the DMU loss with a standard denoising loss, our method ensures that "forgotten" domains are structurally destroyed while the model's core utility and intra-class diversity on retained data remain intact. Empirical evaluations across different benchmarks and models demonstrate that DMU is a versatile solution for large-scale and continuous unlearning requests.

\textbf{Future work:} We will explore the extension of this proxy to the limitation we discussed before and other generative modalities, such as Large Language Models, to further enhance the "right to be forgotten" in the era of foundation models.

{
\small
\bibliographystyle{IEEEtran} 
\bibliography{references}

@article{cooper2024machine,
  title={Machine Unlearning Doesn't Do What You Think: Lessons for Generative AI Policy, Research, and Practice},
  author={Cooper, A Feder and Choquette-Choo, Christopher A and Bogen, Miranda and Jagielski, Matthew and Filippova, Katja and Liu, Ken Ziyu and Chouldechova, Alexandra and Hayes, Jamie and Huang, Yangsibo and Mireshghallah, Niloofar and others},
  journal={arXiv preprint arXiv:2412.06966},
  year={2024}
}

@article{li2025machine,
  title={Machine unlearning: Taxonomy, metrics, applications, challenges, and prospects},
  author={Li, Na and Zhou, Chunyi and Gao, Yansong and Chen, Hui and Zhang, Zhi and Kuang, Boyu and Fu, Anmin},
  journal={IEEE Transactions on Neural Networks and Learning Systems},
  year={2025},
  publisher={IEEE}
}

@article{regulation2016regulation,
  title={Regulation (EU) 2016/679 of the European Parliament and of the Council},
  author={Regulation, Protection},
  journal={Regulation (eu)},
  volume={679},
  number={2016},
  pages={10--13},
  year={2016}
}

@article{li2025sculpting,
  title={Sculpting Memory: Multi-Concept Forgetting in Diffusion Models via Dynamic Mask and Concept-Aware Optimization},
  author={Li, Gen and Xiao, Yang and Ji, Jie and Deng, Kaiyuan and Hui, Bo and Guo, Linke and Ma, Xiaolong},
  journal={arXiv preprint arXiv:2504.09039},
  year={2025}
}

@article{deng2026forget,
  title={Forget Many, Forget Right: Scalable and Precise Concept Unlearning in Diffusion Models},
  author={Deng, Kaiyuan and Li, Gen and Xiao, Yang and Hui, Bo and Ma, Xiaolong},
  journal={arXiv preprint arXiv:2601.06162},
  year={2026}
}

@inproceedings{xiao2025efficient,
  title={Efficient Knowledge Graph Unlearning with Zeroth-order Information},
  author={Xiao, Yang and Ye, Ruimeng and Liu, Bohan and Ma, Xiaolong and Hui, Bo},
  booktitle={Proceedings of the 34th ACM International Conference on Information and Knowledge Management},
  pages={3509--3519},
  year={2025}
}

@article{lee2025continual,
  title={Continual Unlearning for Text-to-Image Diffusion Models: A Regularization Perspective},
  author={Lee, Justin and Mai, Zheda and Yoo, Jinsu and Fan, Chongyu and Zhang, Cheng and Chao, Wei-Lun},
  journal={arXiv preprint arXiv:2511.07970},
  year={2025}
}

@inproceedings{xiao2025knowledge,
  title={Knowledge Graph Unlearning with Schema},
  author={Xiao, Yang and Ye, Ruimeng and Hui, Bo},
  booktitle={Proceedings of the 31st International Conference on Computational Linguistics},
  pages={3541--3546},
  year={2025}
}

@article{kawamura2025approximate,
  title={Approximate Domain Unlearning for Vision-Language Models},
  author={Kawamura, Kodai and Goto, Yuta and Yanagi, Rintaro and Kataoka, Hirokatsu and Irie, Go},
  journal={arXiv preprint arXiv:2510.08132},
  year={2025}
}

@inproceedings{xiao2025right,
  title={The Right to be Forgotten in Pruning: Unveil Machine Unlearning on Sparse Models},
  author={Xiao, Yang and Li, Gen and Ji, Jie and Ye, Ruimeng and Ma, Xiaolong and Hui, Bo},
  booktitle={NeurIPS 2025 Workshop on Regulatable ML}
}

@article{zhang2024unlearncanvas,
  title={Unlearncanvas: Stylized image dataset for enhanced machine unlearning evaluation in diffusion models},
  author={Zhang, Yihua and Fan, Chongyu and Zhang, Yimeng and Yao, Yuguang and Jia, Jinghan and Liu, Jiancheng and Zhang, Gaoyuan and Liu, Gaowen and Kompella, Ramana Rao and Liu, Xiaoming and others},
  journal={arXiv preprint arXiv:2402.11846},
  year={2024}
}

@article{thakral2025continual,
  title={Continual unlearning for foundational text-to-image models without generalization erosion},
  author={Thakral, Kartik and Glaser, Tamar and Hassner, Tal and Vatsa, Mayank and Singh, Richa},
  journal={arXiv preprint arXiv:2503.13769},
  year={2025}
}

@article{deng2026forget2,
  title={Forget-It-All: Multi-Concept Machine Unlearning via Concept-Aware Neuron Masking},
  author={Deng, Kaiyuan and Hui, Bo and Li, Gen and Ji, Jie and Qin, Minghai and Yuan, Geng and Ma, Xiaolong},
  journal={arXiv preprint arXiv:2601.06163},
  year={2026}
}

@inproceedings{liu2022right,
  title={The right to be forgotten in federated learning: An efficient realization with rapid retraining},
  author={Liu, Yi and Xu, Lei and Yuan, Xingliang and Wang, Cong and Li, Bo},
  booktitle={IEEE INFOCOM 2022-IEEE conference on computer communications},
  pages={1749--1758},
  year={2022},
  organization={IEEE}
}

@article{tarun2023fast,
  title={Fast yet effective machine unlearning},
  author={Tarun, Ayush K and Chundawat, Vikram S and Mandal, Murari and Kankanhalli, Mohan},
  journal={IEEE transactions on neural networks and learning systems},
  volume={35},
  number={9},
  pages={13046--13055},
  year={2023},
  publisher={IEEE}
}

@inproceedings{gandikota2023erasing,
  title={Erasing concepts from diffusion models},
  author={Gandikota, Rohit and Materzynska, Joanna and Fiotto-Kaufman, Jaden and Bau, David},
  booktitle={Proceedings of the IEEE/CVF international conference on computer vision},
  pages={2426--2436},
  year={2023}
}

@inproceedings{wu2025erasing,
  title={Erasing undesirable influence in diffusion models},
  author={Wu, Jing and Le, Trung and Hayat, Munawar and Harandi, Mehrtash},
  booktitle={Proceedings of the Computer Vision and Pattern Recognition Conference},
  pages={28263--28273},
  year={2025}
}

@inproceedings{gandikota2024unified,
  title={Unified concept editing in diffusion models},
  author={Gandikota, Rohit and Orgad, Hadas and Belinkov, Yonatan and Materzy{\'n}ska, Joanna and Bau, David},
  booktitle={Proceedings of the IEEE/CVF winter conference on applications of computer vision},
  pages={5111--5120},
  year={2024}
}

@article{fan2023salun,
  title={Salun: Empowering machine unlearning via gradient-based weight saliency in both image classification and generation},
  author={Fan, Chongyu and Liu, Jiancheng and Zhang, Yihua and Wong, Eric and Wei, Dennis and Liu, Sijia},
  journal={arXiv preprint arXiv:2310.12508},
  year={2023}
}

@inproceedings{basak2024forget,
  title={Forget more to learn more: Domain-specific feature unlearning for semi-supervised and unsupervised domain adaptation},
  author={Basak, Hritam and Yin, Zhaozheng},
  booktitle={European Conference on Computer Vision},
  pages={130--148},
  year={2024},
  organization={Springer}
}

@inproceedings{venkateswara2017deep,
  title={Deep hashing network for unsupervised domain adaptation},
  author={Venkateswara, Hemanth and Eusebio, Jose and Chakraborty, Shayok and Panchanathan, Sethuraman},
  booktitle={Proceedings of the IEEE Conference on Computer Vision and Pattern Recognition},
  pages={5018--5027},
  year={2017}
}

@inproceedings{li2017deeper,
  title={Deeper, broader and artier domain generalization},
  author={Li, Da and Yang, Yongxin and Song, Yi-Zhe and Hospedales, Timothy M},
  booktitle={Proceedings of the IEEE international conference on computer vision},
  pages={5542--5550},
  year={2017}
}

@misc{tunstall2023zephyr,
      title={Zephyr: Direct Distillation of LM Alignment}, 
      author={Lewis Tunstall and Edward Beeching and Nathan Lambert and Nazneen Rajani and Kashif Rasul and Younes Belkada and Shengyi Huang and Leandro von Werra and Clémentine Fourrier and Nathan Habib and Nathan Sarrazin and Omar Sanseviero and Alexander M. Rush and Thomas Wolf},
      year={2023},
      eprint={2310.16944},
      archivePrefix={arXiv},
      primaryClass={cs.LG}
}

@article{lan2025diffusion,
  title={Diffusion-Based Cross-Modal Feature Extraction for Multi-Label Classification},
  author={Lan, Tian and Zheng, Yiming and Yin, Jianxin},
  journal={arXiv preprint arXiv:2509.15553},
  year={2025}
}

@article{chang2024sparsedit,
  title={Sparsedit: Token sparsification for efficient diffusion transformer},
  author={Chang, Shuning and Wang, Pichao and Tang, Jiasheng and Wang, Fan and Yang, Yi},
  journal={arXiv preprint arXiv:2412.06028},
  year={2024}
}

@article{lee2025v,
  title={V-Warper: Appearance-Consistent Video Diffusion Personalization via Value Warping},
  author={Lee, Hyunkoo and Jang, Wooseok and Yang, Jini and Kim, Taehwan and Kim, Sangoh and Jung, Sangwon and Kim, Seungryong},
  journal={arXiv preprint arXiv:2512.12375},
  year={2025}
}

@article{pal2025llm,
  title={Llm unlearning reveals a stronger-than-expected coreset effect in current benchmarks},
  author={Pal, Soumyadeep and Wang, Changsheng and Diffenderfer, James and Kailkhura, Bhavya and Liu, Sijia},
  journal={arXiv preprint arXiv:2504.10185},
  year={2025}
}

@article{li2024wmdp,
  title={The wmdp benchmark: Measuring and reducing malicious use with unlearning},
  author={Li, Nathaniel and Pan, Alexander and Gopal, Anjali and Yue, Summer and Berrios, Daniel and Gatti, Alice and Li, Justin D and Dombrowski, Ann-Kathrin and Goel, Shashwat and Phan, Long and others},
  journal={arXiv preprint arXiv:2403.03218},
  year={2024}
}

@inproceedings{lu2024mace,
  title={Mace: Mass concept erasure in diffusion models},
  author={Lu, Shilin and Wang, Zilan and Li, Leyang and Liu, Yanzhu and Kong, Adams Wai-Kin},
  booktitle={Proceedings of the IEEE/CVF Conference on Computer Vision and Pattern Recognition},
  pages={6430--6440},
  year={2024}
}

@article{zhao2024separable,
  title={Separable multi-concept erasure from diffusion models},
  author={Zhao, Mengnan and Zhang, Lihe and Zheng, Tianhang and Kong, Yuqiu and Yin, Baocai},
  journal={arXiv preprint arXiv:2402.05947},
  year={2024}
}

@article{deng2026no,
  title={No Concept Escapes the Audit: Auditing-Aware Unlearning for Verifiable Concept Erasure in Diffusion Models},
  author={Deng, Kaiyuan and Li, Yuchen and Li, Gen and Xiao, Yang and Yuan, Geng and Yuan, Xiaoyong and Hui, Bo and Ma, Xiaolong},
  journal={arXiv preprint arXiv:2610.05401},
  year={2026}
}

@article{deng2026your,
  title={Your Unlearning Gives You Away: Identifying Erased Concepts in Diffusion Models},
  author={Deng, Kaiyuan and Li, Yuchen and Xiao, Yang and Hui, Bo and Yuan, Geng and Ma, Xiaolong},
  journal={arXiv preprint arXiv:2610.05601},
  year={2026}
}

@article{li2026fade,
  title={FADE: Frame-Aware Diffusion-Transformer-based Multi-Concept Erasure for Video Unlearning},
  author={Li, Yuchen and Deng, Kaiyuan and Feng, Chaoran and Tang, Zhenyu and Yuan, Li},
  journal={arXiv preprint arXiv:2610.03980},
  year={2026}
}
}

% \begin{figure*}[t]
%     \centering
%     \begin{subfigure}[t]{0.32\textwidth}
%         \centering
%         \includegraphics[width=\linewidth]{Figures/HA_request.png}
%         \caption{HA under different requests.}
%         \label{fig:ha_request}
%     \end{subfigure}
%     \hfillx
%     \begin{subfigure}[t]{0.32\textwidth}
%         \centering
%         \includegraphics[width=\linewidth]{Figures/UA_request.png}
%         \caption{UA under different requests.}
%         \label{fig:ua_request}
%     \end{subfigure}
%     \hfill
%     \begin{subfigure}[t]{0.32\textwidth}
%         \centering
%         \includegraphics[width=\linewidth]{Figures/RA_request.png}
%         \caption{RA under different requests.}
%         \label{fig:ra_request}
%     \end{subfigure}
%     \caption{Performance comparison under different unlearning requests.}
%     \label{fig:request_comparison}
% \end{figure*}

%%%%%%%%%%%%%%%%%%%%%%%%%%%%%%%%%%%%%%%%%%%%%%%%%%%%%%%%%%%%

\appendix
\section{Detailed Experiment Setups}
Provided from UnlearnCanvas benchmarks~\cite{zhang2024unlearncanvas}, we use the pretrained object and style classifier to get \textbf{UA}, \textbf{RA}$_{pos}$, \textbf{RA}$_{anc}$, and \textbf{StyleAcc}$_{R}$ on UnlearnCanvas dataset. Following their plan, we adopted a ViT-L/16 model pretrained on ImageNet and finetune it on
OfficeHome~\cite{venkateswara2017deep} and PACS~\cite{li2017deeper}. Datasets are split into the train set and test set with a ratio of 9: 1.
After hyper-parameter tuning, the classifiers are trained with the Adam optimizer at a learning rate of 0.01 for 10 epochs. We use \texttt{openai/clip-vit-base-patch32} to compute the \textbf{CLIP}$_{R}$ between generated images and their corresponding prompts. 
\subsection{GPU, Training and Test Latency}
We run all experiments in 4 L40S GPUs. 

\textbf{BU}: Each experiment requires approximately 210 minutes. The entire image generation process typically takes approximately 1 hour to complete. 

\textbf{SU}: Each experiment requires approximately $\text{Number of styles} \times$ 210 minutes. The entire image generation process typically takes approximately 1 hour to complete.

\subsection{How many objects and how many styles?}
\paragraph{UnlearnCanvas} Here we show the specific objects and styles we use in this dataset.

6 Objects: Birds, Trees, Cats, Horses, Jellyfish, Human.

6 Styles: Abstractionism, Byzantine, Cartoon, Cold Warm, Ukiyoe, Van Gogh.

12 Styles: Abstractionism Blossom Season, Byzantine, Cartoon, Cold Warm,  Expressionism, Ukiyoe, Van Gogh, Meta Physics, Sponge Dabbed, Warm Smear, Winter.

15 Styles: Abstractionism, Artist Sketch, Blossom Season, Byzantine, Cartoon, Cold Warm, Dapple, Expressionism, Greenfield, Ukiyoe, Van Gogh, Meta Physics, Sponge Dabbed, Warm Smear, Winter.

25 Styles: Abstractionism, Artist Sketch, Blossom Season, Bricks, Byzantine, Cartoon, Cold Warm, Color Fantasy, Comic Etch, Crayon, Cubism, Dadaism, Dapple, Defoliation, Early Autumn, Expressionism, Fauvism, French, Greenfield, Ukiyoe, Van Gogh, Meta Physics, Sponge Dabbed, Warm Smear, Winter.

30 Styles: Abstractionism, Artist Sketch, Blossom Season, Bricks, Byzantine, Cartoon, Cold Warm, Color Fantasy, Comic Etch, Crayon, Cubism, Dadaism, Dapple, Defoliation, Early Autumn, Expressionism, Fauvism, French, Greenfield, Glowing Sunset, Gorgeous Love, Impressionism, Ink Art, Joy, Ukiyoe, Van Gogh, Meta Physics, Sponge Dabbed, Warm Smear, Winter.

\paragraph{OfficeHome} Here we show the specific objects and styles we use in this dataset.

6 Objects: Bike, Calculator, Monitor, Refrigerator, Telephone, TV.

2 Styles: Art, Clipart.
\paragraph{PACS} Here we show the specific objects and styles we use in this dataset.

7 Objects: All Objects in PACS.

3 Styles: art painting, cartoon, sketch.

\subsection{Image Generation}
\textbf{Random Seeds.} For the positive and negative sets, we use seven random seeds: 42, 144, 79, 27, 599, 123, and 577. For the anchor set, we use five random seeds: 188, 288, 388, 488, and 588.

\textbf{Generation Protocol.} For each domain (i.e., each object--style pair) and each seed, we generate 4 images using the relevant seed + (0-3), because we have 4 GPUs. The final test dataset is obtained by aggregating samples across all seeds. 

\textbf{Weight of DMU (Lambda).} 
We conduct an ablation study on the loss weighting parameter $\lambda$ using the UnlearnCanvas dataset, focusing on the Birds class with the Stable Diffusion v1.5 backbone. When $\lambda$ = 0.0, the DMU objective is  removed, reducing the training to standard fine-tuning on the retain set (i.e., positive and anchor samples only). As shown in Table~\ref{tab:abla_lambdas}, this setting leads to significantly degraded unlearning performance, particularly in terms of UA and HA, indicating that the DMU mechanism is essential for effective forgetting. In contrast, across a wide range of $\lambda$ values from 0.05 to 1.00, the model consistently achieves strong unlearning results with high UA and HA scores, while maintaining stable retention performance (RA$_{pos}$ and RA$_{anc}$) and generation quality (StyleAcc$_{R}$ and CLIP$_{R}$). These results demonstrate that the proposed method is robust to the choice of $\lambda$, and that the unlearning effectiveness is not sensitive to precise tuning of this hyperparameter. For the experiments presented in the main text, we initially selected 0.15 as the weight via random sampling; given the significant results shown, a default weight of 0.15 was subsequently applied to all DMU experiments within the main text. However, this part indicated that varying the weight within the range of 0.05 to 1.00 yielded no substantial differences.

\begin{table}[]
    \centering
    \caption{Ablation Studies on Different Lambdas}
    \begin{tabular}{c|>{\columncolor{HAcolor}}c|>{\columncolor{UAcolor}}c|c|c|c|c}
    \toprule
    \midrule
        $\lambda$ & HA (↑) & UA (↑) & RA$_{pos}$(↑) & RA$_{anc}$ (↑) & StyleAcc$_{R}$(↑) & CLIP$_{R}$(↑) \\
    \midrule
        0.00& 47.42& 25.79& 72.42& 93.59& 92.27&29.15 \\
    \midrule
        0.05& 86.26& 89.58& 80.64& 89.16& 95.51&29.32\\
        0.15& 90.45& 94.59& 84.29& 93.17& 94.99&29.16\\
        0.20& 82.60& 79.17& 83.82& 85.04& 94.14&29.33\\
        0.25& 90.58& 95.83& 89.93& 86.48& 93.08&29.32\\
        0.30& 87.15& 89.58& 87.28& 84.73& 93.92&29.33\\
        0.50& 82.46& 77.08& 86.71& 84.21& 93.31&29.28\\
        0.75& 88.08& 97.92& 84.68& 83.08& 91.03&29.26\\
        1.00& 86.90& 95.83& 82.37& 83.70& 90.57&29.09\\
    \midrule
    \bottomrule
    \end{tabular}
    \label{tab:abla_lambdas}
\end{table}

\section{Case Study}
Here we show some cases in Figure~\ref{fig:case_birds}:
\begin{figure}
    \centering
    \includegraphics[width=1.0\linewidth]{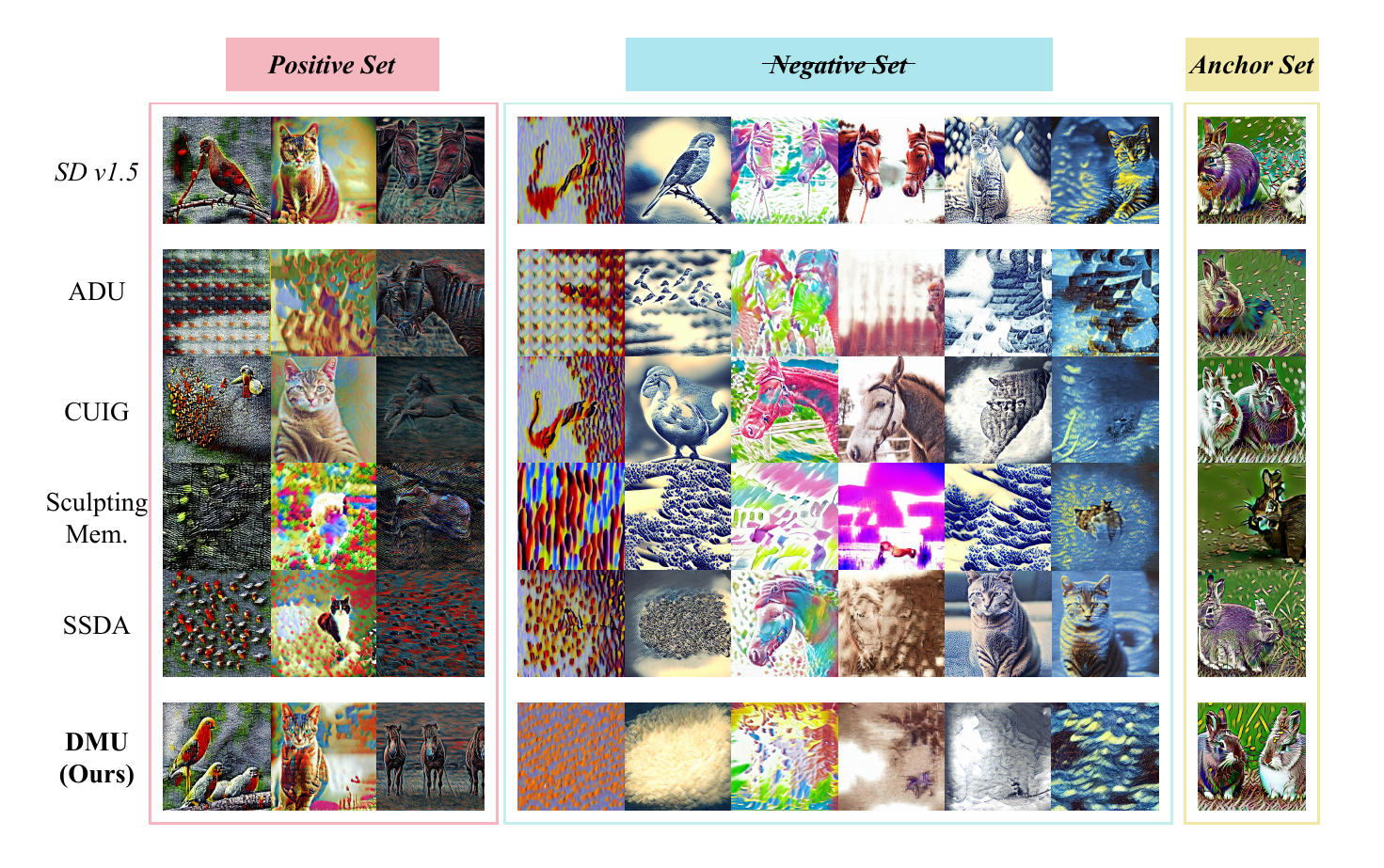}
    \caption{Case }
    \label{fig:case_birds}
\end{figure}
\section{Application in LLM Unlearning.}

\begin{table}[]
    \centering
    \caption{Application in LLM Unlearning}
    \begin{tabular}{l|c|c|c}
    \toprule
    \midrule
        Model & WMDP-Bio (↓) & WMDP-Cyber (↓) & MMLU (↑) \\
        
    \midrule
        ZEPHYR-7B & 63.7 & 44.0 & 58.1 \\
        + NPO & 53.7 & 34.6 & 50.4 \\
        + GA & 51.8 & 30.4 & 45.9 \\
        + RMU & 27.3 & 26.7 & 56.4 \\
        + RMU-LAT & 31.7 & 29.6 & 57.5 \\
        + DMU & 29.3 & 27.3 & 56.6 \\
    \midrule
    \bottomrule
    \end{tabular}
    \label{tab:llm_dmu}
\end{table}
We further extend DMU to the large language model setting based on Zephyr-7B~\cite{tunstall2023zephyr}. For the retain set, we first construct a set of reference points by randomly partitioning the tokens in the first batch into 500 groups and computing the mean hidden state at layer l for each group. These reference points are initialized via a short warmup stage of w optimization steps. During unlearning, for each forget token, we identify its k nearest reference points in the layer-l representation space, compute a softmax over their distances, and use the resulting weighted centroid as the target representation. The forget loss is then defined similarly to RMU~\cite{pal2025llm}, except that the target is no longer a random vector but this adaptive centroid. Meanwhile, the reference points are continuously updated using an exponential moving average (EMA) based on retain tokens to ensure stability. As shown in Table~\ref{tab:llm_dmu}, DMU achieves strong unlearning performance on both WMDP-Bio~\cite{li2024wmdp} and WMDP-Cyber~\cite{li2024wmdp} while maintaining competitive general capability on MMLU. Compared with prior methods such as NPO, GA, and RMU variants, our approach provides a better trade-off between effective forgetting and knowledge retention, demonstrating its applicability beyond diffusion models.

\newpage
\end{document}